%% file: main.tex
\documentclass{article}

\usepackage{microtype}
\usepackage{graphicx}
\usepackage{subfigure}
\usepackage{booktabs} 

\usepackage{hyperref}

\usepackage[accepted]{icml2024}

\usepackage{amsmath,bm}
\usepackage{amssymb}
\usepackage{mathtools}
\usepackage{amsthm}
\usepackage{pifont}

\usepackage{enumitem}
\usepackage{multirow}
\usepackage{makecell}
\usepackage{arydshln} 

\usepackage[capitalize,noabbrev]{cleveref}

\theoremstyle{plain}

\theoremstyle{definition}

\theoremstyle{remark}

\usepackage[textsize=tiny]{todonotes}

\usepackage{comment}
\usepackage{xcolor}

\newcommand{\nickname}{OSN}

\newcommand{\ie}{\textit{i}.\textit{e}.}
\newcommand{\eg}{\textit{e}.\textit{g}.}
\newcommand{\etc}{\textit{etc}.}

\newcommand{\wrt}{\textit{w}.\textit{r}.\textit{t}}

\icmltitlerunning{OSN: Infinite Representations of Dynamic 3D Scenes from Monocular Videos}

\begin{document}

\twocolumn[
\icmltitle{OSN: Infinite Representations of Dynamic 3D Scenes from Monocular Videos}



\icmlsetsymbol{equal}{*}

\begin{icmlauthorlist}
\icmlauthor{Ziyang Song}{polyusz,vlar}
\icmlauthor{Jinxi Li}{polyusz,vlar}
\icmlauthor{Bo Yang}{polyusz,vlar}
\end{icmlauthorlist}

\icmlaffiliation{polyusz}{Shenzhen Research Institute, The Hong Kong Polytechnic University, Shenzhen, China\\ \phantom{xx}}
\icmlaffiliation{vlar}{vLAR Group, The Hong Kong Polytechnic University, Hung Hom, HKSAR \\ \phantom{xx}}
\icmlcorrespondingauthor{Bo Yang}{bo.yang@polyu.edu.hk}

\icmlkeywords{Machine Learning, ICML}

\vskip 0.3in
]



\printAffiliationsAndNotice{}  

\begin{abstract}
It has long been challenging to recover the underlying dynamic 3D scene representations from a monocular RGB video. Existing works formulate this problem into finding a single most plausible solution by adding various constraints such as depth priors and strong geometry constraints, ignoring the fact that there could be infinitely many 3D scene representations corresponding to a single dynamic video. In this paper, we aim to learn all plausible 3D scene configurations that match the input video, instead of just inferring a specific one.  To achieve this ambitious goal, we introduce a new framework, called \textbf{\nickname{}}. The key to our approach is a simple yet innovative object scale network together with a joint optimization module to learn an accurate scale range for every dynamic 3D object. This allows us to sample as many faithful 3D scene configurations as possible. Extensive experiments show that our method surpasses all baselines and achieves superior accuracy in dynamic novel view synthesis on multiple synthetic and real-world datasets. Most notably, our method demonstrates a clear advantage in learning fine-grained 3D scene geometry. Our code and data are available at \href{https://github.com/vLAR-group/OSN}{https://github.com/vLAR-group/OSN}
\end{abstract}

\section{Introduction}
\input{chaps/01_intro}

\section{Related Works}
\input{chaps/02_liter}

\section{\nickname{}}
\input{chaps/03_meth}

\section{Experiments}
\input{chaps/04_exp}

\section{Conclusion}
\input{chaps/05_sum}

\section*{Acknowledgements}
This work was supported in part by National Natural Science Foundation of China (62271431), in part by Shenzhen Science and Technology Innovation Commission (JCYJ20210324120603011),
in part by Research Grants Council of Hong Kong (25207822 \& 15225522).

\section*{Impact Statement}
This paper presents work whose goal is to advance the field of Machine Learning. There are many potential societal consequences of our work, none which we feel must be specifically highlighted here.

\clearpage
\bibliography{references}
\bibliographystyle{icml2024}

\clearpage
\appendix
\input{chaps/06_app}



\end{document}

%% file: chaps/01_intro.tex
Our 3D world is primarily a collection of many movable objects, often rigid with dynamics over time, \eg{}, balloons flying in the air and balls bouncing back and forth. It has long been desired to recover object structures of such dynamic 3D scenes just from casually captured monocular videos \cite{Costeira1995,Fitzgibbon2000}. However, this problem is highly ill-posed, as there could be infinitely many geometric explanations that match an input dynamic video. 

To tackle the challenge of modelling a dynamic 3D scene from a single video, existing methods formulate this problem into finding a singular, most plausible solution by adding various additional constraints, such as monocular depths \cite{Cai2022,Song2023,Zhao2024}, camera poses \cite{Li2021c,Du2021a,Stier2023}, shape templates \cite{Weng2022}, consistency \cite{Gao2021,Li2023b,Guo2023,Yang2024}, physical restrictions \cite{Yang2023b}, diffusion priors \cite{Tu2023,Wang2024}, \etc{}. Thanks to the powerful SDF \cite{Park2019}, NeRF \cite{Mildenhall2020}, and Gaussian Splatting \cite{Kerbl2023} as backbones, these methods demonstrate excellent performance in modelling dynamic 3D scenes including dynamic novel view rendering and 3D shape reconstruction. However, such a formulation oversimplifies the problem of monocular-based dynamic 3D modelling.
\begin{figure}[t]
    \centering
    \includegraphics[width=1.0\columnwidth]{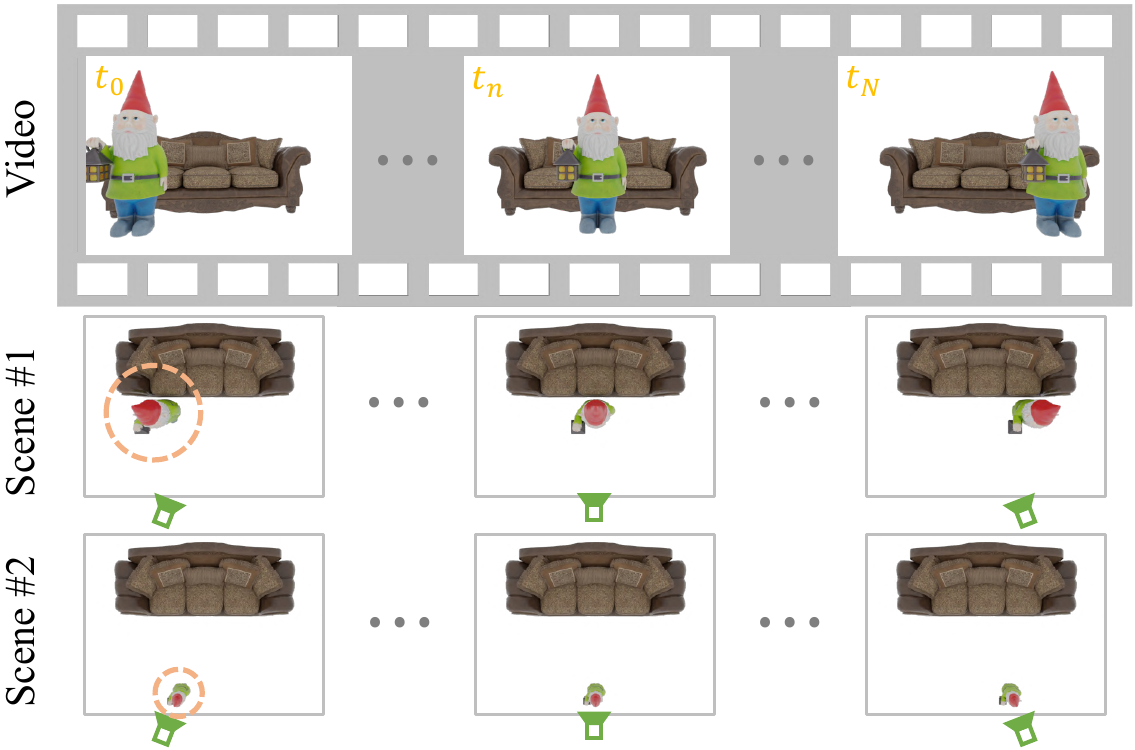}
    \vskip -0.15in
    \caption{An illustration of multiple correct 3D scene configurations that match the same dynamic monocular video.} 
    \label{fig:task_illustration}
    \vskip -0.3in
\end{figure}

As illustrated in Figure \ref{fig:task_illustration}, regarding a dynamic monocular video with a moving object inside (shown in the top row), clearly, there are numerous 3D scene configurations matched with the video. For example, both Scene\#1 with a large object (the middle row) and Scene\#2 with a small object (the bottom row) are true solutions. This means that the problem of monocular-based dynamic 3D modelling inherently needs to be resolved by inferring all correct solutions, instead of just estimating a specific one as has been done in all existing works.
\begin{figure*}[t]
    \centering
    \includegraphics[width=2.0\columnwidth]{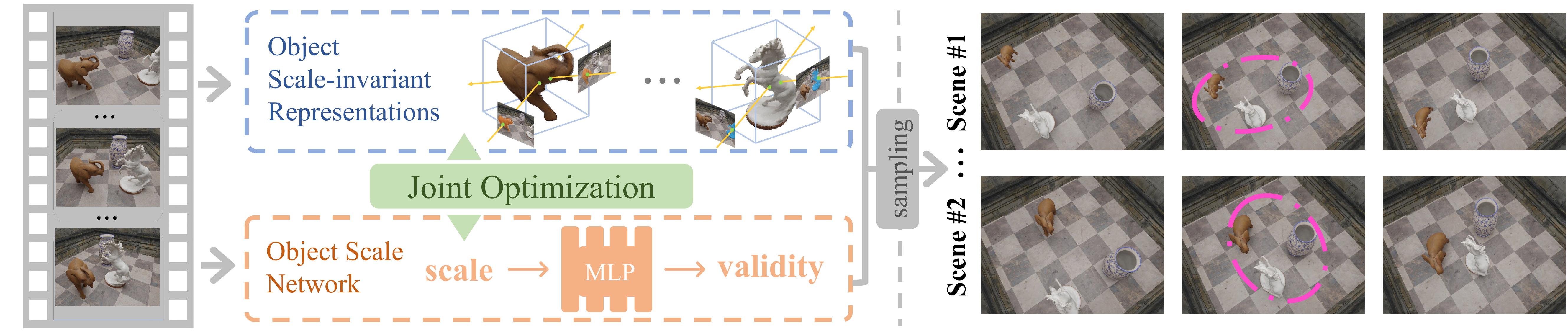}
    \vskip -0.15in
    \caption{An illustration of our framework. Given a dynamic video as input, our Object Scale-invariant Representation module (the blue block) and the Object Scale Network (the orange block) aim to represent all faithful 3D scene representations, allowing infinitely sampling of different 3D scenes (the rightmost block) after they are jointly optimized. Circles highlight the differences between the two scenes.}
    \label{fig:overview}
    \vskip -0.15in
\end{figure*}

With this motivation, we aim to address dynamic 3D modelling by learning all faithful 3D scene representations just from a monocular RGB video without depth scans and camera poses. However, this relaxed and ambitious problem is particularly challenging, as it is so far unclear how to learn or even represent infinite correct 3D scene configurations. 

Upon closer investigation, we find that the core difficulty of this new problem lies in how to infer an accurate scale range for each dynamic object to perfectly match the input video. After that, all faithful 3D scene configurations can be simply obtained by combining different scaled 3D objects. However, object scales are tightly compounded with object-camera joint motions as already illustrated in Figure \ref{fig:task_illustration}. More importantly, the relative scales between multiple objects are also coupled to each other due to mutual visual occlusions captured in the input video. These make the object scale learning extremely hard.

In this paper, we focus on modelling dynamic 3D scenes with rigid objects, leaving deformable dynamic reconstruction for future exploration. In particular, given a monocular video capturing complex dynamics of multiple rigid objects by a moving camera, our task is to learn the scale ranges of all dynamic objects along with recovering per-object shape and appearance. Ultimately, all 3D scene configurations that match the input video can be comprehensively recovered, allowing dynamic novel view synthesis at any timestamp on any specific 3D scene configuration.

As shown in Figure \ref{fig:overview}, we introduce a new framework with three major components: 1) an object scale-invariant representation module (blue block), 2) an object scale network (orange block), and 3) a joint optimization module. For the first component, it is flexible to adopt an existing 3D object representation network such as SDF \cite{Park2019}, NeRF \cite{Mildenhall2020}, or the recent Gaussian Splatting \cite{Kerbl2023}. This module only aims to learn per-object shape and appearance representations within the same 3D unit volume, \ie{}, scale-invariant. 

The \textbf{object scale network} is the core of our framework, aiming at learning relative scale ranges of all dynamic 3D objects, whereas the third module involves carefully designed \textit{scaled composite rendering} and \textit{soft Z-buffer rendering} algorithms, driving both object scale-invariant representations and object scale ranges to be accurately learned. Once the per-object shape, appearance, and scale ranges are well learned, our framework allows infinitely sampling of faithful and valid 3D scene configurations, as illustrated in the rightmost block of Figure \ref{fig:overview}. Our framework is named \textbf{\nickname{}} and our contributions are:
\begin{itemize}[leftmargin=*]
\setlength{\itemsep}{1pt}
\setlength{\parsep}{1pt}
\setlength{\parskip}{1pt}
\vspace{-0.45cm}
    \item We introduce the first framework to represent dynamic 3D scenes in infinitely many ways that match an input monocular video, while existing methods only learn a single solution with additional priors.
    \item We design an object scale network with scaled composite rendering and soft Z-buffer rendering techniques to jointly learn scale ranges of all dynamic objects, allowing dynamic 3D scenes to be comprehensively represented.
    \item We demonstrate superior results in dynamic novel view synthesis on multiple 3D datasets. In addition, our method shows great authenticity in 3D scene sampling.
\end{itemize}

%% file: chaps/02_liter.tex
\begin{figure*}[t]
    \centering
    \includegraphics[width=1.8\columnwidth]{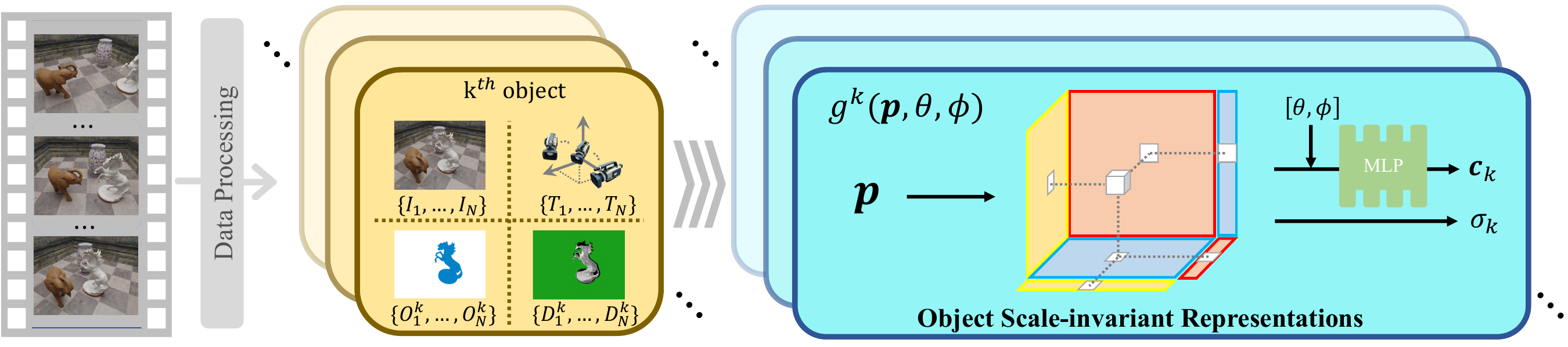}
    \vskip -0.15in
    \caption{The yellow block shows that the input video will first be preprocessed into per-object information. After that, the shape and appearance of each dynamic object will be separately represented by a scale-invariant TensoRF model as shown by the light blue block.}
    \label{fig:network}
    \vskip -0.2in
\end{figure*}

\textbf{Static 3D Representations}: Static 3D objects and scenes can be represented by voxels \cite{Chan2016}, point clouds \cite{Fan2017}, octrees \cite{Tatarchenko2017}, meshes \cite{Kato2017}, and primitives \cite{Zou2017}. However, these representations are usually limited by the spatial resolution and high memory cost. Recent implicit representations demonstrate excellent performance in novel view rendering and shape reconstruction, including occupancy fields (OF) \cite{Mescheder2019,Chen2019g}, (un)signed distance fields (U/SDF) \cite{Park2019,Chibane2020a}, and radiance fields (NeRF) \cite{Mildenhall2020}. Since these implicit representations take 3D points as input, it is usually time-consuming to render novel views or explicitly regress 3D surfaces. To overcome this limitation, the very recent 3D Gaussian Splatting \cite{Kerbl2023} and RayDF \cite{Liu2023a} directly learn to represent 3D surfaces, achieving real-time rendering speed. In our framework, the first object scale-invariant module is amenable to existing 3D representations or their variants. 

\textbf{Dynamic 3D Representations}: Recent advances in dynamic 3D representations mostly extend existing static 3D representations such as SDF, NeRF, and Gaussian Splatting by adding the time dimension \textit{t} as an additional input. They either disentangle the dynamic scenes into a canonical template and a time-dependent motion field \cite{Pumarola2021,Barron2021a,Park2021,Tretschk2021,Cai2022,Fang2022}, or directly model a space-time radiance field \cite{Xian2021,Li2022,Park2023}, or adopt a combination of both \cite{Li2021c,Gao2021,Du2021a,Liu2023}. These methods are mostly designed for dynamic 3D scenes with deformable objects/scenes or specific categories.  

For dynamic scenes with rigid moving objects, existing works usually disentangle each dynamic object into a canonical space with a time-dependent $SE(3)$ pose \cite{Yuan2021,Kundu2022,Song2023,Xie2023}. Thanks to the simplicity, our first object scale-invariant module also follows this strategy.

\textbf{Constrained} \textit{vs} \textbf{Unconstrained 3D Modelling}: Recovering 3D shapes and appearances just from 2D images is usually an ill-posed problem due to the lack of sufficient geometry constraints \cite{Hartley2004}. For the particular challenging case of monocular-based dynamic 3D modelling, existing works turn it into a tightly constrained problem by adding additional priors or restrictions, including the availability of monocular depths \cite{Yoon2020,Du2021a,Cai2022,Song2023,Zhao2024}, accurate camera poses \cite{Li2021c,Stier2023,Wang2023,Park2024}, shape templates \cite{Weng2022}, temporal consistency \cite{Gao2021,Barron2021a,Guo2023,Tian2023,Choe2023,Wang2023a,Liu2023,Li2023b}, physical restrictions \cite{Yang2023b}, diffusion priors \cite{Tu2023,Wang2024}, \etc{}. By doing so, these methods achieve remarkable performance in estimating just a single and most plausible 3D representation. However, in this paper, we tackle a relaxed and unconstrained problem, aiming at learning all faithful 3D scene representations from a single video. 

\textbf{Difference from 3D Generative Models}: 3D generative models aim to learn the distributions of observed 3D datasets by generating conditional or unconditional new 3D objects or scenes. Thanks to the sophisticated deep generative models such as VAEs \cite{Kingma2014}, GANs \cite{Goodfellow2014}, diffusion models \cite{Ho2020}, and the recent large language models (LLMs) \cite{Brown2020,Radford2021,Zhao2023}, 3D generative models show remarkable progress as comprehensively discussed in recent surveys \cite{Shi2022,Li2023d}. In this paper, our method can recover infinitely many 3D scene configurations by learning the valid ranges of all dynamic objects. Nevertheless, all the learned 3D scene representations are just derived from a single data point (the input monocular video), instead of large 3D datasets.

%% file: chaps/03_meth.tex
\subsection{Preliminary}\label{sec:method_preliminary}
Given a monocular RGB video with $N$ frames (timestamps) $\{ I_1, ..., I_N \}$, there are $K$ rigid objects inside moving in different directions. Note that, the actual entire static background is simply regarded as one of the $K$ dynamic objects, as all objects are moving regarding the mobile camera. 

As shown in Figure \ref{fig:network}, our framework consists of a data preprocessing stage followed by other components. In the preprocessing stage, we firstly segment all $K$ objects in the frames by a pretrained SAM \cite{Kirillov2023}, followed by per-object tracking and per-pixel optical flow estimation using pretrained TAM \cite{Yang2023} and RAFT \cite{Teed2020}, obtaining pixel-level associated $T$ masks for any $k^{th}$ object, \ie{}, $\{ O_1^k, ..., O_N^k \}$. And then, for the $k^{th}$ object, we simply use SfM \cite{Schonberger2016} to estimate the camera-to-$k^{th}$-object relative poses at each frame (denoted as $\{ T_1^k, ..., T_N^k \}$), as well as the $k^{th}$-object-to-camera relative depth values at corresponding pixels obtained by SfM triangulation (denoted as $\{ D_1^k, ..., D_N^k \}$). Here are three points to be clarified:
\begin{itemize}[leftmargin=*]
\setlength{\itemsep}{1pt}
\setlength{\parsep}{1pt}
\setlength{\parskip}{1pt}
\vspace{-0.3cm}
    \item Both the poses and depth values can only be estimated for each object, and the scales cannot be shared across multiple objects in the same scene, fundamentally because the motion and scale of each object are visually compounded with the unknown camera motion.
    \item In the preprocessing stage, our framework is also flexible to many simple and classic alternatives such as key pixel matching by SIFT \cite{Lowe2004} followed by motion clustering \cite{Elhamifar2013} and SfM. We opt for pretrained models thanks to their excellent performance.
    \item Such a preprocessing stage itself cannot alleviate the difficulty of our unconstrained problem, as it lacks additional geometry priors to restrict the infinitely many solutions.   
\end{itemize}\vspace{-0.3cm}

Having the preprocessed per-object information (RGB, object masks, relative camera poses, relative depths), we simply regard each object as static by treating its total $N$ timestamp RGB frames as multi-view images, and then use a single network to represent each object respectively. In particular, we adopt an existing TensoRF \cite{Chen2022b} to represent each object, though other variants can also be used. The total $K$ TensoRF models form our \textbf{object scale-invariant representation module}, as illustrated by the light blue block of Figure \ref{fig:network}. More implementation details are in Appendix. This module together with our core \textbf{object scale network} will be trained by the \textbf{joint optimization module}. Detailed designs are discussed in Sections \ref{sec:method_osn}\&\ref{sec:method_jointOpt}.

\subsection{Object Scale Network}\label{sec:method_osn}
Since there is no existing work to learn valid scale ranges of dynamic objects just from a monocular video, a na\"ive idea is to directly feed the $N$ images into a network to regress $2K$ parameters, representing the lower and upper bounds of $K$ object scales. However, such a network can hardly be optimized, essentially due to the lack of ground truth object scale ranges as supervision signals. When we design a network, two key factors need to be considered:
\begin{itemize}[leftmargin=*]
\setlength{\itemsep}{1pt}
\setlength{\parsep}{1pt}
\setlength{\parskip}{1pt}
\vspace{-0.3cm}
    \item The scales of multiple dynamic objects are always relative and intertwined with each other. This means that we can randomly pick up an object as an anchor and set its scale to be 1, while only estimating the valid scale ranges of the remaining $K-1$ objects in the scene.
    \item Learning multi-object scale ranges is actually a binary classification problem from a monocular video. We can only verify whether the learned or sampled object scales are valid by comparing them with the input video. 
\end{itemize}\vspace{-0.3cm}

To this end, we design a conceptually simple object scale network, which only consists of MLPs. It takes a sampled multi-object scale combination as input and predicts a validity score between $0\sim 1$, where 1 represents the input scale combination valid and 0 otherwise. 
\begin{figure}[h]\vskip -0.05in
    \centering
    \includegraphics[width=1\columnwidth]{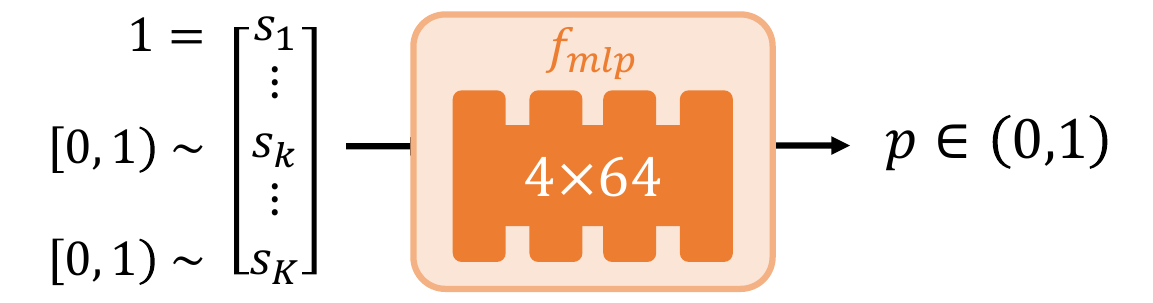}
    \vskip -0.15in
    \caption{An illustration of our object scale network.}
    \label{fig:method_osn}
    \vskip -0.2in
\end{figure}

As shown in Figure \ref{fig:method_osn}, for the total $K$ objects in a 3D scene, we select one object (usually the largest object such as the background for simplicity) and set its scale as 1, \ie{}, $s_1 = 1$, and the remaining $K-1$ object scales $\{s_2, \cdots, s_K\}$ are uniformly sampled from a predefined normalized range $[0, 1)$. Note that, an unbounded/unnormalized range sampling would pose difficulties to optimize in practice. To map these normalized object ranges back to the 3D scene volume, we simply apply the following linear operation:
\begin{equation}
    \bar{s}_k = \frac{\mathcal{D}_{near}^{scene}}{\mathcal{D}^k_{near}}(1 - s_k) + \frac{\mathcal{D}_{far}^{scene}}{\mathcal{D}^k_{far}}s_k
\end{equation} \vskip -0.15in
\textit{where} $\mathcal{D}_{near}^k$ and $\mathcal{D}_{far}^k$ are the near/far distances of the $k^{th}$ object in its own 3D object volume along $z$ axis. The $\mathcal{D}_{near}^{scene}$ and $\mathcal{D}_{far}^{scene}$ are predefined near/far distances of the 3D scene volume along $z$ axis. Naturally, we choose $\mathcal{D}_{near}^{scene}$ to be smaller than all $\mathcal{D}_{near}^k$, while $\mathcal{D}_{far}^{scene}$ to be larger than all $\mathcal{D}_{far}^k$. For simplicity, here we just use the camera center of the first video frame as both the 3D scene coordinate and the $k^{th}$ 3D object coordinate. Both $\mathcal{D}_{near}^k$ and $\mathcal{D}_{far}^k$ can be easily chosen for the $k^{th}$ object based on its sparse point cloud obtained by SfM in the data preprocessing stage in Section \ref{sec:method_preliminary}. 
More implementation details of the network are in Appendix. 

To sum up, our object scale network basically learns the validity score $p$ for every (sampled) normalized $K$ scales as: $p = f_{mlp}([s_1,\cdots, s_K])$. Subsequently, the denormalized $K$ object scales $[\bar{s}_1,\cdots, \bar{s}_K]$ can be calculated. Now, the key issue is how to effectively optimize the network (Section \ref{sec:method_jointOpt}), so that given any multi-object scale samples during testing, it can predict correct scores, and ultimately we can recover all valid scale ranges of $K$ objects.  

\subsection{Joint Optimization}\label{sec:method_jointOpt}
After the data preprocessing stage (Section \ref{sec:method_preliminary}), for any $k^{th}$ object, we have its object masks $\{O^k_1, \cdots, O^k_N\}$, masked images $\{I_1*O^k_1, \cdots, I_N*O^k_N\}$, the corresponding camera-to-$k^{th}$-object poses $\{T_1^k, \cdots, T_N^k\}$, and the $k^{th}$-object-to-camera relative depths $\{D^k_1, \cdots, D^k_N\}$ as all training signals. Now we need to optimize the $K$ object scale-invariant representation networks and the object scale network. 

As to any $k^{th}$ object scale-invariant representation network, \ie{}, a TensoRF model denoted as $g^k(\mathbf{p}, \theta, \phi)$ where $\mathbf{p}/\theta/\phi$ are any query point and angle, it can be easily optimized independently based on standard volume rendering using an RGB loss $\ell_{rgb}^k$, optionally with a depth loss $\ell_{depth}^k$ \cite{Deng2022} as shown below, where both are $\ell_2$ losses. \vskip -0.2in
\begin{equation}\label{eq:meth_independent_loss}
    g^k \xleftarrow{\text{optimize}} (\ell_{rgb}^k + \ell_{depth}^k)
\end{equation} \vskip -0.15in

Nevertheless, such a separate training scheme tends to be inferior, as it fails to take into account the mutual visual occlusions caused by other objects. Most importantly, the object scale network $f_{mlp}$ has yet to be optimized, and it can only be optimized by composing all scaled $K$ objects. In this regard, we propose the following two techniques: scaled composite rendering and soft Z-buffer rendering. 

\textbf{Scaled Composite Rendering}: Having the $K$ object scale-invariant representations, \ie{}, $\{g^1, \cdots, g^K\}$, and a sampled multi-object scale combination $[\bar{s}_1, \cdots, \bar{s}_K]$, our scaled composite rendering aims to render images by combining the shape, color, and scale information of all $K$ objects, so to minimize the discrepancy with all input video frames.

In particular, for any specific ray (pixel) selected from the $n^{th}$ image ($I_n$) of the input video, we can easily identify which object this pixel belongs to, according to our preprocessed data. Assuming it belongs to the $k^{th}$ object, we then sample $M$ points $[\mathbf{p}_1^k, \cdots, \mathbf{p}_M^k]$ along that ray, denoted as $\mathbf{r}^k$ with viewing angle $[\theta^k, \phi^k]$ calculated from the camera-to-$k^{th}$-object poses. Clearly, we can first obtain the corresponding colors and densities from the $k^{th}$ object representation $g^k$:\vskip -0.3in
\begin{equation}
    \big\{[\mathbf{c}_1^k, \cdots, \mathbf{c}_M^k], [\mathbf{\sigma}_1^k, \cdots, \mathbf{\sigma}_M^k]\big\}  \leftarrow g^k([\mathbf{p}_1^k, \cdots, \mathbf{p}_M^k], \theta^k, \phi^k)
\end{equation} \vskip -0.1in

Next, we need to obtain the colors and densities of these $M$ points after transforming them to another object representation space such as the $\hat{k}^{th}$ object space. Here, we need to take into account the camera poses $\{T_n^k; T_n^{\hat{k}}\}$ at the current $n^{th}$ image for both objects, and their object scales $\{\bar{s}_k; \bar{s}_{\hat{k}}\}$. Specifically, the $M$ points are scaled and then transformed from the $k^{th}$ object space to the $\hat{k}^{th}$ object space as follows:
\vskip -0.2in
\begin{equation}\label{eq:meth_point_transform}
    \mathbf{p}_m^{\hat{k}} = T_n^{\hat{k}}(T_n^{k})^{-1} \circ \big(\frac{\bar{s}_k\mathbf{p}_m^{k}}{\bar{s}_{\hat{k}}}\big)
\end{equation} \vskip -0.1in
Note that, for the object scales $\{\bar{s}_k; \bar{s}_{\hat{k}}\}$ used in Equation \ref{eq:meth_point_transform}, their corresponding sampled scales $[s_1,\cdots, s_K]$ should be deemed as valid, meaning that the estimated validity score $p = f_{mlp}([s_1,\cdots, s_K])$ should be larger than a threshold, \eg{}, 0.95 in our implementation. Otherwise, we need to resample until the estimated score is above 0.95, even though the network $f_{mlp}$ is not fully optimized in the early stage.

Similarly, the viewing angle is also transformed from the $k^{th}$ object space to the $\hat{k}^{th}$ object space as follows:
\vskip -0.2in
\begin{equation}\label{eq:meth_view_transform}
    [\theta^{\hat{k}}, \phi^{\hat{k}}] = T_n^{\hat{k}}(T_n^{k})^{-1} \circ [\theta^k, \phi^k]
\end{equation} \vskip -0.1in
Naturally, we can obtain the colors and densities of the transformed $M$ points at the $\hat{k}^{th}$ object space as follows:
\begin{equation}
    \big\{[\mathbf{c}_1^{\hat{k}}, \cdots, \mathbf{c}_M^{\hat{k}}], [\mathbf{\sigma}_1^{\hat{k}}, \cdots, \mathbf{\sigma}_M^{\hat{k}}]\big\}  \leftarrow g^{\hat{k}}([\mathbf{p}_1^{\hat{k}}, \cdots, \mathbf{p}_M^{\hat{k}}], \theta^{\hat{k}}, \phi^{\hat{k}})
\end{equation} \vskip -0.1in
In this way, for this ray $\mathbf{r}^k$ and the sampled $M$ points $[\mathbf{p}_1^k, \cdots, \mathbf{p}_M^k]$ along it, we can obtain their colors and densities in all $K$ object spaces as shown below:
\vskip -0.3in
\begin{equation}
\begin{cases}
\text{$1^{st}$ object $g^1$:} & [\mathbf{c}_1^1, \cdots, \mathbf{c}_M^1], [\mathbf{\sigma}_1^1, \cdots, \mathbf{\sigma}_M^1] \\
 \cdots \\
\text{$K^{th}$ object $g^K$:} & [\mathbf{c}_1^K, \cdots, \mathbf{c}_M^K], [\mathbf{\sigma}_1^K, \cdots, \mathbf{\sigma}_M^K] 
\end{cases}       
\end{equation}\vskip -0.1in
Lastly, we combine colors and densities from all $K$ object spaces using an existing composite volume rendering in Total-Recon \cite{Song2023}, generating the final color $\mathbf{c}(\mathbf{r}^k)$, depth value $d(\mathbf{r}^k)$, and object segmentation $\mathbf{o}(\mathbf{r}^k)$ (a soft one-hot vector) for the selected ray $\mathbf{r}^k$. 

The whole object scale-invariant representation module can be optimized by the following composite scene-level RGB loss, optionally with the depth loss and object segmentation loss (cross-entropy). Note that, the ground truth depth value $\bar{d}(\mathbf{r}^k)$ (retrieved from $D_n^k$ of the $k^{th}$ object) needs to be scaled because the predicted depth is a composite of scaled 3D points. The ground truth object segmentation $\mathbf{\bar{o}}(\mathbf{r}^k)$ per ray is also retrieved from our preprocessed data $O_n^k$. 
\vskip -0.3in
\begin{align}
\ell_{rgb}^{scene} &= \sum_{\mathbf{r}^k}||\mathbf{c}(\mathbf{r}^k) - \mathbf{\bar{c}}(\mathbf{r}^k)|| \label{eq:meth_scaled_CR_rgb}  \\
\ell_{depth}^{scene} &= \sum_{\mathbf{r}^k}||d(\mathbf{r}^k) - s_k*\bar{d}(\mathbf{r}^k)|| \label{eq:meth_scaled_CR_depth} \\
\ell_{seg}^{scene} &= \sum_{\mathbf{r}^k}CE\Big(\mathbf{o}(\mathbf{r}^k), \mathbf{\bar{o}}(\mathbf{r}^k)\Big) \label{eq:meth_scaled_CR_seg}
\end{align}\vskip -0.2in
Nevertheless, the above three losses are unable to optimize our object scale network $f_{mlp}$, as the scales used in Equation \ref{eq:meth_point_transform} are essentially sampled values, not estimated by $f_{mlp}$.  

\textbf{Soft Z-buffer Rendering}:
To optimize our object scale network $f_{mlp}$, we need to obtain supervision signals, \ie{}, ground truth validity scores, for a sufficient number of sampled object scales, covering both valid and invalid scale combinations. A na\"ive strategy is to extensively use our scaled composite rendering. 

For example, for the selected ray $\mathbf{r}^k$, we randomly sample as many as $H$ different combinations of scales $\{\mathcal{S}^1, \cdots, \mathcal{S}^H\}$:
\vskip -0.3in
\begin{equation}\label{eq:meth_soft_z_scalesSamp}
\Big\{\mathcal{S}^1= [s_1^1, \cdots, s_K^1], 
 \cdots, 
\mathcal{S}^H= [s_1^H, \cdots, s_K^H] 
\Big\}
\end{equation}\vskip -0.1in
Then, we use our scaled composite rendering to obtain all corresponding candidate object segmentation results:
\vskip -0.2in
\begin{equation}
\Big\{\mathbf{o}^1(\mathbf{r}^k), 
 \cdots, 
\mathbf{o}^H(\mathbf{r}^k)
\Big\}
\end{equation}\vskip -0.1in
By comparing with the ground truth segmentation $\mathbf{\bar{o}}(\mathbf{r}^k)$, we can easily obtain a pseudo ground validity score $\bar{p}^h$ for each sampled scale combination $\mathcal{S}^h$. In particular,
\vskip -0.2in
\begin{equation}\label{eq:meth_soft_z_validity_label}
\bar{p}^h = \sum \Big(|\mathbf{o}^h(\mathbf{r}^k)|* \mathbf{\bar{o}}(\mathbf{r}^k)\Big) \rightarrow 0/1 
\end{equation}\vskip -0.1in
\textit{where} $|\mathbf{o}^h(\mathbf{r}^k)|$ means we use \textit{argmax} to convert the soft one-hot vector into a hard one. In this way, we can assign 0/1 labels for all sampled $H$ scale combinations:
\vskip -0.2in
\begin{equation}
\{\mathcal{S}^1, \cdots, \mathcal{S}^H\} \xleftarrow{\text{assign labels}} \{\bar{p}^1, \cdots, \bar{p}^H\} 
\end{equation}\vskip -0.1in
So far, we can use these labels to optimize our object scale network $f_{mlp}$ by the binary cross-entropy loss:
\vskip -0.2in
\begin{equation}\label{eq:meth_soft_z_bce}
f_{mlp} \xleftarrow{\text{optimize}} \ell_{bce} = \sum_{\mathbf{r}^k} \Big( \sum_h BCE(p^h, \bar{p}^h) \Big)
\end{equation}\vskip -0.1in
Ultimately, this loss function drives the object scale network $f_{mlp}$ to predict correct validity scores for every sampled scale combination, such that all (scaled) 3D objects will satisfy the mutual visual occlusions at all training images. 

However, recall that, for every selected ray $\mathbf{r}^k$, we need to sample a large number of $H$ scale combinations, followed by $H$ times of scaled composite rendering to obtain object segmentation. This is extremely time-consuming. To tackle this, we introduce a fast soft Z-buffer rendering strategy to obtain object segmentation.

In particular, for the selected ray $\mathbf{r}^k$ at the $n^{th}$ image ($I_n$) of the input video, we first transform its viewing angle $[\theta^k, \phi^k]$ to all other object spaces using Equation \ref{eq:meth_view_transform}, getting:
\vskip -0.2in
\begin{equation}\label{eq:meth_soft_z_rays}
\Big\{\cdots,[\theta^{\hat{k}}, \phi^{\hat{k}}], \cdots, [\theta^K, \phi^K] \Big\}
\end{equation}\vskip -0.1in
Then, in each object space, we uniformly sample $M$ points at fixed positions along the transformed ray, \eg{}, $[\theta^{\hat{k}}, \phi^{\hat{k}}]$ of the $\hat{k}^{th}$ object space, followed by depth rendering, same as a regular and independent NeRF model does. In this way, we get $K$ depth values (up-to-scale) along the $K$ transformed rays (Equation \ref{eq:meth_soft_z_rays}) in the total $K$ object spaces, respectively:
\vskip -0.2in
\begin{equation}\label{eq:meth_soft_z_independent_depth}
\Big\{\cdot\cdot,[\theta^{\hat{k}}, \phi^{\hat{k}}], \cdot\cdot, [\theta^K, \phi^K] \Big\} \xrightarrow[\text{rendering}]{\text{independent}} \Big\{\cdot\cdot, d^{\hat{k}}(\mathbf{r}^k), \cdot\cdot, d^K(\mathbf{r}^k) \Big\} 
\end{equation}\vskip -0.1in
Inspired by the Z-buffer algorithm in rasterization-based rendering, for the selected ray $\mathbf{r}^k$, we determine the object segmentation of each sampled scale combination $\mathcal{S}^h$ (Equation \ref{eq:meth_soft_z_scalesSamp})
according to the order of scaled depth values of the total $K$ objects. In particular, for the sampled scale $\mathcal{S}^h = [s^h_1, \cdots, s^h_K]$, and the independently rendered depths in Equation \ref{eq:meth_soft_z_independent_depth}, the corresponding object segmentation is computed as:
\vskip -0.3in
\begin{equation}\label{eq:meth_soft_z_objSeg}
\mathbf{o}^h(\mathbf{r}^k) = \Bigg[ \cdots,\frac{e^{-\big(s_k^h*d^{k}(\mathbf{r}^k) \big)}}{
\sum_{k=1}^K e^{-\big(s_k^h*d^{k}(\mathbf{r}^k) \big)}
},\cdots \Bigg]  
\end{equation}\vskip -0.1in
Lastly, we use the same Equation \ref{eq:meth_soft_z_validity_label} to create validity labels, and Equation \ref{eq:meth_soft_z_bce} to optimize the object scale network.

Overall, to get object segmentation results of $H$ sampled scale combinations for a specific ray $\mathbf{r}^k$, the scaled composite rendering needs to query all $K$ object scale-invariant networks $H$ times, while the soft Z-buffer rendering only needs to query once followed by $H$ times of a simple operation in Equation \ref{eq:meth_soft_z_objSeg}, being nearly $H$ times faster overall. 

Note that, the soft Z-buffer rendering can also be used for computing scene-level losses in Equations \ref{eq:meth_scaled_CR_rgb}/\ref{eq:meth_scaled_CR_depth}/\ref{eq:meth_scaled_CR_seg}. Nevertheless, to optimize object scale-invariant networks, for each selected ray $\mathbf{r}^k$, we just need a single (valid) sample of scale combination $\mathcal{S}$, thus the time cost of scaled composite rendering and soft Z-buffer rendering is the same.

\textbf{Joint Training Procedure}: From our design, we can see that optimizing the object scale network relies on relatively reasonable depth results from the object scale-invariant representations. Otherwise, the computed pseudo ground truth validity is unreliable. In return, a reasonably good object scale network can also benefit the optimization of object scale-invariant representations. To this end, we follow an intuitively simple two-stage training procedure. 
\begin{itemize}[leftmargin=*]
\setlength{\itemsep}{1pt}
\setlength{\parsep}{1pt}
\setlength{\parskip}{1pt}
\item \textbf{Stage 1 - Boostrapping Per-object Representations}: In the very beginning, we apply independent volume rendering for each object, separately optimizing per-object TensoRF model by the RGB/depth losses in Equation \ref{eq:meth_independent_loss}. 
\item \textbf{Stage 2 - Alternative Optimization}: We alternatively optimize, 1) the object scale network by:
\vskip -0.15in
\begin{equation}
f_{mlp} \xleftarrow{\text{optimize}} \ell_{bce} 
\end{equation}\vskip -0.1in
2) all object scale-invariant representation networks by: 
\vskip -0.2in
\begin{equation}
\{g^1, \cdots, g^K\} \xleftarrow{\text{optimize}} (\ell_{rgb}^{scene} + \ell_{depth}^{scene} + \ell_{seg}^{scene}) 
\end{equation}\vskip -0.1in
After $R$ rounds of alternation, the object scale-invariant networks and the object scale network can be effectively trained. More implementation details are in Appendix.

\end{itemize}

%% file: chaps/04_exp.tex
\begin{figure*}[h]
    \centering    
    \includegraphics[width=2.08\columnwidth]{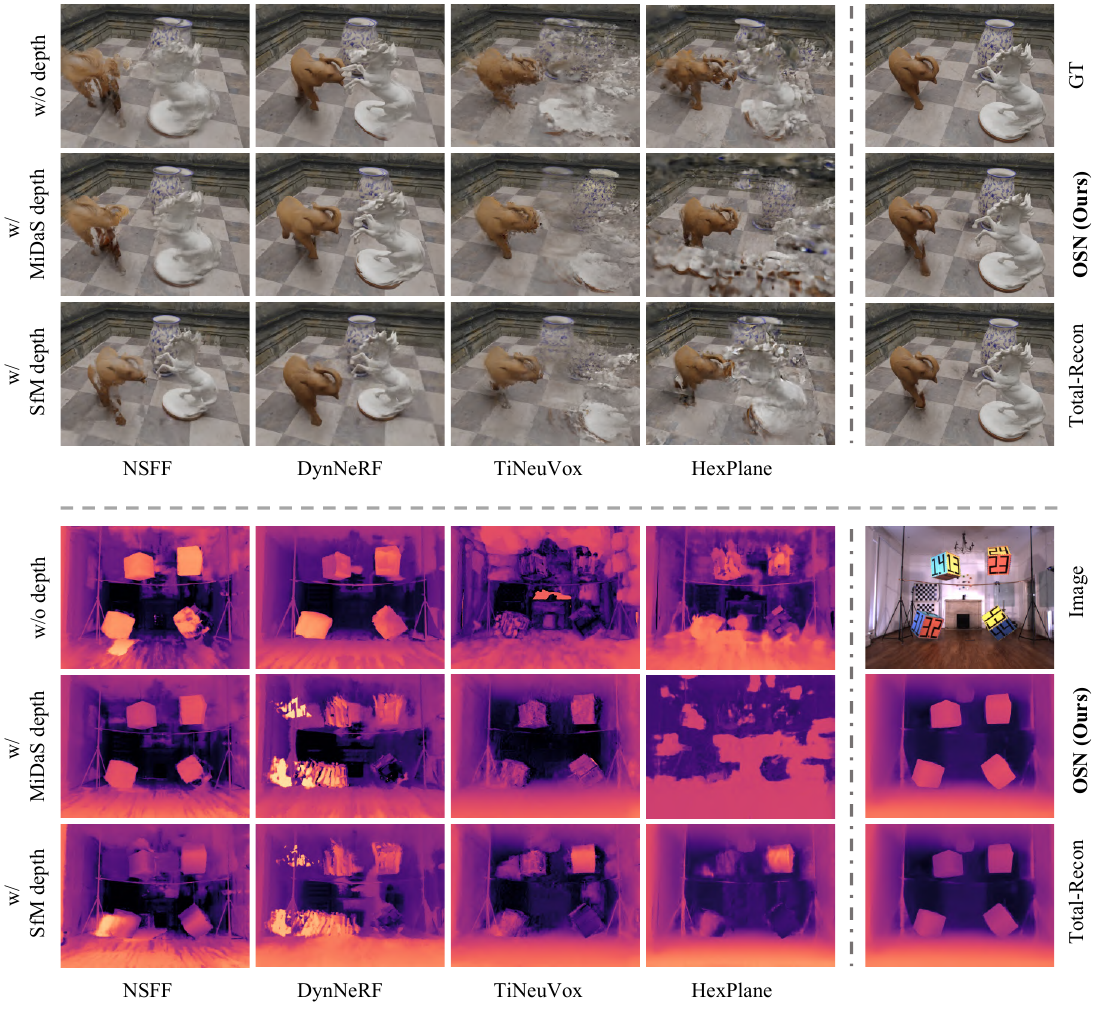}
    \vskip -0.18in
    \caption{Qualitative results of dynamic novel view RGB/depth synthesis on the Dynamic Indoor Scene and Oxford Multimotion Datasets.}
    \label{fig:qual_res}
    \vskip -0.05in
\end{figure*}

\begin{table*}[t]
\caption{Quantitative results of all methods for dynamic novel view synthesis on three datasets. The methods are trained with different depth supervision: 1) w/o depth, 2) w/ MiDaS depth, and 3) w/ per-object SfM depth.}
\tabcolsep= 0.2cm 
\centering
\begin{scriptsize}
\begin{tabular}{cl|cccc|ccc|ccc}
    \hline
    \multirow{2}{*}{} & & \multicolumn{4}{c}{Dynamic Indoor Scene Dataset } & \multicolumn{3}{c}{Oxford Multimotion Dataset} & \multicolumn{3}{c}{NVIDIA Dynamic Scene Dataset} \\
    Depth Sup. & Method & PSNR$\uparrow$ & SSIM$\uparrow$ & LPIPS$\downarrow$ & SSIMAE$\downarrow$ & PSNR$\uparrow$ & SSIM$\uparrow$ & LPIPS$\downarrow$ & PSNR$\uparrow$ & SSIM$\uparrow$ & LPIPS$\downarrow$ \\
    \hline
    \multirow{4}{*}{\makecell[c]{1)}} & NSFF\cite{Li2021c} & 21.428 & 0.720 & 0.313 & 0.378 & 16.687 & 0.616 & 0.249 & 21.766 & 0.669 & 0.229 \\
    & DynNeRF\cite{Gao2021} & 21.479 & 0.752 & 0.277 & 0.417 & 16.858 & 0.627 & 0.244 & 25.705 & 0.827 & 0.117 \\
    & TiNeuVox\cite{Fang2022} & 21.705 & 0.655 & 0.306 & 0.484 & 16.433 & 0.613 & 0.325 & 22.922 & 0.618 & 0.262 \\
    & HexPlane\cite{Cao2023a} & 18.637 & 0.581 & 0.480 & 0.962 & 17.084 & 0.631 & 0.221 & 20.169 & 0.555 & 0.286 \\
    \hline
    \multirow{4}{*}{\makecell[c]{2)}} & NSFF\cite{Li2021c} & 20.900 & 0.698 & 0.349 & 0.494 & 17.094 & 0.623 & 0.244 & 27.459 & 0.861 & 0.075 \\
    & DynNeRF\cite{Gao2021} & 22.272 & 0.767 & 0.257 & 0.309 & 16.521 & 0.622 & 0.259 & \underline{29.452} & \textbf{0.895} & \underline{0.054} \\
    & TiNeuVox\cite{Fang2022} & 23.288 & 0.698 & 0.269 & 0.329 & \underline{18.508} & 0.668 & \underline{0.197} & 23.029 & 0.621 & 0.193 \\
    & HexPlane\cite{Cao2023a} & 17.968 & 0.528 & 0.535 & 1.395 & 15.843 & 0.576 & 0.338 & 19.312 & 0.471 & 0.334 \\
    \hline
    \multirow{5}{*}{\makecell[c]{3)}} & NSFF\cite{Li2021c} & 21.280 & 0.684 & 0.347 & 0.467 & 17.093 & 0.616 & 0.245 & 23.733 & 0.733 & 0.194 \\
    & DynNeRF\cite{Gao2021} & 21.421 & 0.742 & 0.296 & 0.509 & 16.786 & 0.624 & 0.281 & 24.498 & 0.771 & 0.176 \\
    & TiNeuVox\cite{Fang2022} & 22.197 & 0.685 & 0.285 & 0.368 & 18.043 & \underline{0.670} & 0.208 & 22.691 & 0.591 & 0.215 \\
    & HexPlane\cite{Cao2023a} & 20.217 & 0.623 & 0.373 & 0.458 & 17.137 & 0.631 & 0.203 & 23.220 & 0.720 & 0.150 \\
    & \textbf{\nickname{}(Ours)} & \textbf{25.984} & \textbf{0.861} & \textbf{0.115} & \textbf{0.094} & \textbf{19.671} & \textbf{0.695} & \textbf{0.155} & \textbf{29.588} & \underline{0.892} & \textbf{0.053} \\
    \hline
    2)+3) & Total-Recon\cite{Song2023} & \underline{24.695} & \underline{0.841} & \underline{0.128} & \underline{0.137} & 18.331 & 0.655 & \underline{0.173} & 27.822 & 0.880 & 0.059 \\
    \hline
\end{tabular}\vspace{-0.2cm}
\end{scriptsize}
\label{tab:exp_dnvs}
\end{table*}

\begin{table*}[t]
\caption{Quantitative results of all methods for dynamic novel view synthesis on synthetic ``Gnome House" scene with 50 different ground truth scale combinations. The average performance along with standard deviations on 50 groups of ground truths are reported. The methods are trained with different depth supervision: 1) w/o depth, 2) w/ MiDaS depth, and 3) w/ per-object SfM depth.}
\tabcolsep= 0.25cm 
\centering
\begin{scriptsize}
\begin{tabular}{cl|cccc}
    \hline
    \multirow{2}{*}{} & & \multicolumn{4}{c}{50 Ground Truth Scenes of Gnome House} \\
    Depth Sup. & Method & PSNR$\uparrow$ & SSIM$\uparrow$ & LPIPS$\downarrow$ & SSIMAE$\downarrow$ \\
    \hline
    \multirow{4}{*}{\makecell[c]{1)}} & NSFF\cite{Li2021c} & \underline{19.088$\pm$1.514} & 0.636$\pm$0.026 & 0.385$\pm$0.029 & 0.559$\pm$0.183 \\
    & DynNeRF\cite{Gao2021} & 18.846$\pm$1.227 & 0.645$\pm$0.023 & 0.380$\pm$0.027 & \underline{0.540$\pm$0.156} \\
    & TiNeuVox\cite{Fang2022} & 18.361$\pm$1.159 & 0.539$\pm$0.026 & 0.414$\pm$0.033 & 0.600$\pm$0.140 \\
    & HexPlane\cite{Cao2023a} & 16.762$\pm$0.130 & 0.420$\pm$0.002 & 0.708$\pm$0.005 & 1.688$\pm$0.098 \\
    \hline
    \multirow{4}{*}{\makecell[c]{2)}} & NSFF\cite{Li2021c} & 18.993$\pm$1.485 & 0.592$\pm$0.024 & 0.465$\pm$0.027 & 0.582$\pm$0.180 \\
    & DynNeRF\cite{Gao2021} & 18.759$\pm$1.398 & 0.639$\pm$0.029 & 0.378$\pm$0.032 & 0.579$\pm$0.194 \\
    & TiNeuVox\cite{Fang2022} & 18.978$\pm$1.249 & 0.560$\pm$0.028 & 0.394$\pm$0.035 & 0.619$\pm$0.159 \\
    & HexPlane\cite{Cao2023a} & 17.325$\pm$0.605 & 0.434$\pm$0.015 & 0.626$\pm$0.019 & 1.993$\pm$0.119 \\
    \hline
    \multirow{5}{*}{\makecell[c]{3)}} & NSFF\cite{Li2021c} & 18.214$\pm$0.948 & 0.492$\pm$0.016 & 0.536$\pm$0.020 & 0.776$\pm$0.137 \\
    & DynNeRF\cite{Gao2021} & 18.767$\pm$1.270 & 0.639$\pm$0.026 & 0.382$\pm$0.029 & 0.554$\pm$0.160 \\
    & TiNeuVox\cite{Fang2022} & 18.776$\pm$1.155 & 0.556$\pm$0.027 & 0.396$\pm$0.033 & 0.553$\pm$0.154 \\
    & HexPlane\cite{Cao2023a} & 18.464$\pm$0.767 & 0.492$\pm$0.019 & 0.480$\pm$0.025 & 0.660$\pm$0.130 \\
    & \textbf{\nickname{}(Ours)} & \textbf{22.940$\pm$1.004} & \textbf{0.784$\pm$0.022} & \textbf{0.160$\pm$0.021} & \textbf{0.125$\pm$0.078} \\
    \hline
    2)+3) & Total-Recon\cite{Song2023} & 18.768$\pm$1.535 & \underline{0.666$\pm$0.032} & \underline{0.295$\pm$0.046} & 0.612$\pm$0.212 \\
    \hline
\end{tabular}\vspace{-0.2cm}
\end{scriptsize}
\label{tab:exp_multigt}
\end{table*}

\textbf{Datasets}: Our method is primarily evaluated on three public datasets: 1) an adapted version of the synthetic \textbf{Dynamic Indoor Scene Dataset} \cite{Li2023c} with 4 scenes and each scene has 3$\sim$4 objects with different rigid motions captured, 2) the real-world \textbf{Oxford Multimotion Dataset} \cite{Judd2019} with 4 scenes selected and each scene contains 2$\sim$4 rigid dynamic objects, and 3) the popular but relatively simple real-world \textbf{NVIDIA Dynamic Scene Dataset} \cite{Yoon2020} with 3 scenes selected (deformable scenes excluded) as each scene has only one moving object. More details of datasets are in Appendix.

\textbf{Baselines}: Since there is no prior work estimating dynamic object scales and recovering infinitely many 3D scene representations, we turn to choose the recent single solution based methods as baselines including: 1) \textbf{NSFF} \cite{Li2021c} and 2) \textbf{DynNeRF} \cite{Gao2021} designed for monocular videos, 3) \textbf{TiNeuVox} \cite{Fang2022} and 4) \textbf{Hexplane} \cite{Cao2023a} designed for general dynamic scene modelling with powerful tri-plane architecture, and 5) \textbf{Total-Recon} \cite{Song2023} designed for dynamic scenes with rigid objects.

\textbf{Metrics}: The standard \textbf{PSNR}, \textbf{SSIM}, and \textbf{LPIPS} scores are reported for novel-view RGB synthesis. On the synthetic Dynamic Indoor Scene Dataset, we also report the Scale- and Shift-Invariant Mean Absolute Error (\textbf{SSIMAE}) \cite{Ranftl2022} for novel-view depth synthesis, where a linear transformation is applied to align rendered depth and ground truth before error calculation. For our method, the object segmentation of novel views is also evaluated by Panoptic Quality (\textbf{PQ}) and mean Intersection over Union (\textbf{mIoU}). As pointed out by DyCheck \cite{Gao2022}, some regions in test views may not be observed in the monocular video. We therefore mask out these regions for a fair comparison.  

\textbf{Settings}: Thanks to our new formulation of dynamic 3D modelling, our \nickname{} framework allows us to sample as infinitely many scale combinations as possible to render novel views, to greedily match the single solution provided in datasets. We report our best scores from 1000 samples, though more can be sampled given computation. Unfortunately, all other baselines can only recover one solution of a dynamic 3D scene to calculate their metrics. 

For a fair and extensive comparison, baselines are trained with three different settings of depth supervision: 
\begin{itemize}[leftmargin=*]
\setlength{\itemsep}{1pt}
\setlength{\parsep}{1pt}
\setlength{\parskip}{1pt}
\vspace{-0.3cm}
\item 1) without depth supervision;
\item 2) with depth supervision (scales inherently aligned across multi-view) from a pretrained monocular depth estimator MiDaS \cite{Ranftl2022};
\item 3) with depth supervision from per-object SfM (scales not aligned) which is the same as ours. \vspace{-0.3cm}
\end{itemize}
Note that, during training, NSFF \cite{Li2021c}, DynNeRF \cite{Gao2021}, and Total-Recon \cite{Song2023} make use of the same preprocessed multi-object segmentation results as our method, and the third group experiments are the fairest comparison. For Total-Recon\cite{Song2023}, we leverage the estimated scene depths from MiDaS to align the scale of objects after per-object SfM. Therefore, Total-recon is trained with depth supervision from both MiDaS and per-object SfM. 

\subsection{Evaluation of Dynamic Novel View Synthesis}
We evaluate dynamic novel view synthesis of our \nickname{} and 5 baselines on 3 datasets in 3 settings, with (4$\times$3+1+1)$\times$11 = 154 models trained in a scene-specific fashion.

\begin{figure*}[t]
    \centering    \includegraphics[width=2.0\columnwidth]{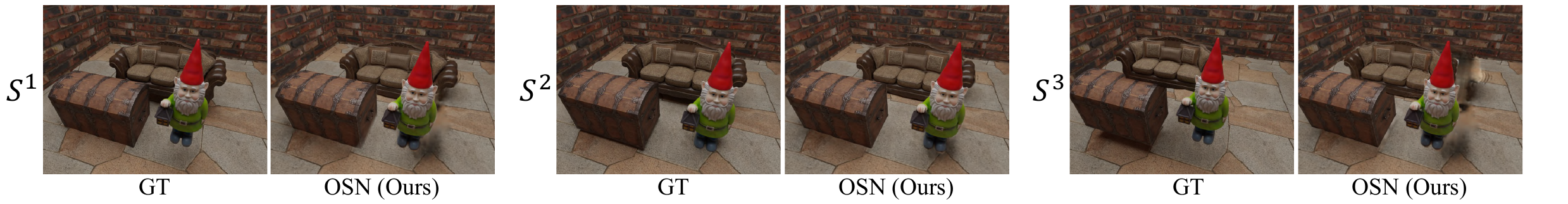}
    \vskip -0.15in
    \caption{Qualitative results of dynamic novel view synthesis for three ground truth 3D scene configurations.}
    \label{fig:exp_res_multigt}
    \vskip -0.01in
\end{figure*}

\begin{figure*}[h]
    \centering    
    \includegraphics[width=2.0\columnwidth]{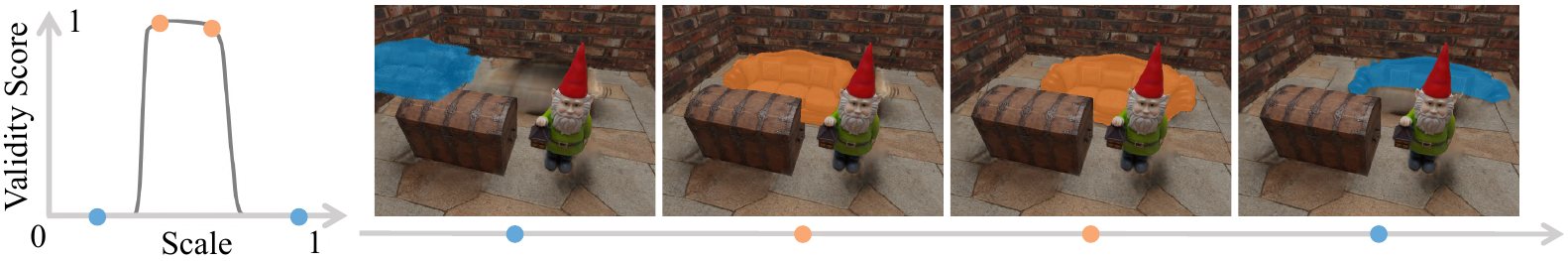}
    \vskip -0.15in
    \caption{An illustration of the learned object scale ranges. We sample 4 scales for \textit{Sofa}, two invalid shown by blue dots and two valid by orange dots. The synthesized novel views demonstrate the learned scales are accurate and visual occlusions well preserved.}
    \label{fig:exp_res_scales}
    \vskip -0.15in
\end{figure*}

\textbf{Analysis}: Table \ref{tab:exp_dnvs} and Figure \ref{fig:qual_res} show the quantitative and qualitative results. We can see that: 
1) Our \nickname{} clearly surpasses all baselines in dynamic novel view RGB synthesis on all datasets, including the extremely strong baselines with pretrained depth priors to learn the most plausible 3D scene representations. In fact, adding pretrained depth priors may incur unreliable geometry constraints and temporal inconsistency. 
2) Total-recon outperforms most of the other baselines, as its representation fully leverages the object rigidity priors. However, it still lags behind our method due to the sub-optimal solution of object scales determined by MiDaS.
3) Most notably, our method achieves superior accuracy in novel view depth estimation with the lowest SSIMAE score of $0.094$ on the Dynamic Indoor Scene Dataset, demonstrating a clear advantage in learning truly faithful 3D geometry thanks to our object scale network which gives explicit but flexible geometry information.

\subsection{Evaluation with Multiple Ground Truths}
Essentially, our framework aims to estimate many correct solutions, and an ideal evaluation benchmark dataset should also have many groups of ground truth images, because there are many ground truth geometric explanations corresponding to the monocular video. In this regard, we aim to truly evaluate dynamic monocular reconstruction as a multi-solution problem. In particular, we take a specific scene ``Gnome House" of the Dynamic Indoor Scene Dataset and create 50 groups of ground truth testing views in Blender. Each group has 210 novel views, and corresponds to a meaningful but different 3D scene configuration where 3D mesh models are applied with different scales, but these 50 scene configurations all correspond to the same set of training views (the input monocular video).

When calculating metric scores for our method and all baselines, we treat the 50 groups of ground truths independently. The most similar predictions out of our 1000 samples will be matched to each group of ground truths, while the baselines always provide the same prediction for all ground truths.

\textbf{Analysis}: As shown in Table \ref{tab:exp_multigt}, our \nickname{} significantly outperforms all baselines regarding both the average performance and variance, since our method can easily produce an approximate solution (out of 1000 samples) for any group of ground truths, while baselines always provide the same solution regardless of different ground truth 3D scene configurations behind the monocular video. Figure \ref{fig:exp_res_multigt} qualitatively demonstrates our \nickname{} produces different solutions for different ground truths.

\subsection{Analysis of Validity Scores}
To further analyze the object scale ranges learned by our \nickname{}, we take ``Gnome House" as an example. As shown in the left plot of Figure \ref{fig:exp_res_scales}, we visualize the learned valid scales of \textit{Sofa} while leaving the other two object scales fixed. By sampling 4 different scales for \textit{Sofa} (two invalid represented by blue dots, two valid by orange dots), we render 4 novel views for the 4 sampled 3D scenes as shown in Figure \ref{fig:exp_res_scales}. We can see that the \textit{Sofa} appears out of the floor and then disappears, correctly occluded by the front two objects at the two valid samples. This means that our object scale network indeed learns the visual relationships and captures authentic scale ranges.

\subsection{Ablation Study}
Since our object scale network only consists of simple 4-layer MLPs, we mainly ablate our joint optimization module on the Dynamic Indoor Scene Dataset.

\textbf{(1) Removing Bootstrapping}: This is to evaluate how the early independent per-object representation learning would help the latter overall optimization of our framework. 

\textbf{(2) Different Rounds of Alternative Optimization}: In default, the alternation round is set as 5, \ie{}, $R=5$. 
\begin{table}[h]\vspace{-0.5cm}
\tabcolsep= 0.12cm 
\caption{Ablation study on the joint optimization module.}
\centering
\begin{scriptsize}
\begin{tabular}{l|cccccc}
    \hline
    & PSNR$\uparrow$ & SSIM$\uparrow$ & LPIPS$\downarrow$ & SSIMAE$\downarrow$ & PQ$\uparrow$ & mIoU$\uparrow$ \\
    \hline
    w/o bootstrapping & 24.467 & 0.831 & 0.135 & 0.174 & 89.139 & 89.258 \\
    $R=1$ & 23.123 & 0.810 & 0.165 & 0.266 & 75.558 & 78.896 \\
    $R=5$ \textbf{(\nickname{})} & \textbf{25.984} & \textbf{0.861} & \textbf{0.115} & \textbf{0.094} & \textbf{92.211} & \textbf{92.451} \\
    $R=10$ & 25.065 & 0.836 & 0.127 & 0.108 & 90.610 & 90.610 \\
    \hline
\end{tabular}
\end{scriptsize}\vspace{-0.5cm}
\label{tab:exp_res_ablation}
\end{table}

Table \ref{tab:exp_res_ablation} shows the ablation results. It can be seen that: 1) Optimizing only one round is not sufficient for the object scale network and the object scale-invariant networks to benefit each other. 2) The early independent per-object optimization is indeed helpful, but our framework would not collapse without it. 3) Excessively training many rounds may not be necessary, as incorrect scale combinations may slip in and lead to inferior object representations over time.    

%% file: chaps/05_sum.tex
In this paper, we demonstrate for the first time that dynamic 3D scenes should be and can be represented in infinitely many ways from a monocular RGB video. This is achieved by a novel object scale network together with a joint optimization module to truly learn a valid scale range for each dynamic object. Extensive experiments validate the effectiveness of our approach on synthetic and real-world datasets with multiple dynamic objects of complex dynamics. We hope that our new formulation of the challenging monocular-based dynamic 3D scene modelling could open up new opportunities for the field of study. Our future work will focus on more challenging deformable 3D scenes.

%% file: chaps/06_app.tex
The appendix includes:
\begin{itemize}[leftmargin=*] \vspace{-0.4cm}
\setlength{\itemsep}{1pt}
\setlength{\parsep}{1pt}
\setlength{\parskip}{1pt}
    \item Discussion on Application Scenarios.
    \item Details of Network.
    \item Details of the Joint Optimization.
    \item Details of Datasets.
    \item Evaluation of Scale Estimation.
    \item Evaluation of Invariance to Object Orders.
    \item Analysis on Number of Possible Ground Truths and Sampled Solutions.
    \item Evaluation on Large and Variable Number of Objects
    \item Evaluation on Self-Driving Scenes.
    \item Evaluation on Daily Scenes.
    \item More Quantitative \& Qualitative Results.
    \vspace{-0.2cm}
\end{itemize}

\section{Discussion on Application Scenarios}
\label{sec:app_scene}

The problem of monocular-based dynamic 3D scene reconstruction can be generally divided into two situations: 
\begin{itemize}[leftmargin=*]
\setlength{\itemsep}{1pt}
\setlength{\parsep}{1pt}
\setlength{\parskip}{1pt}
\vspace{-0.3cm}
\item Situation \#1: Alongside the monocular frames, additional sensors (\eg{}, depth scanner) or prior knowledge (\eg{}, object-ground contact) are also available for the algorithm. Common examples include \textit{autonomous driving, robot manipulation, pedestrian walking,} \etc{}. 
\item Situation \#2: In many other daily scenarios where videos are casually captured by a single camera on a mobile phone, and the captured dynamic objects may not always touch the ground. Common examples include \textit{basketballs bouncing in the air, balloons flying, paragliding, motocross jumping, scenes of Oxford Multimotion dataset,} \etc{}.    
\vspace{-0.3cm}
\end{itemize}

In Situation \#1, the relative scales of multiple dynamic objects can be determined by finding a singular, most plausible solution, which has been extensively studied in literature. However, in Situation \#2, due to the lack of additional constraints, there could be many geometric explanations for the input dynamic monocular video. For example, an object could be interpreted as a small object near the camera or a large one far away, given the same up-to-scale relative motions between objects and the camera. Our \nickname{} is primarily designed for Situation \#2, though it can be easily adapted for Situation \#1 where we just need to sample a specific scale combination (given or estimated from addition constraints).

\section{Details of Networks}

\subsection{Object Scale-Invariant Representation Module}
We adopt VM decomposition in TensoRF \cite{Chen2022b} to parametrize the volumetric radiance field of each rigid object. The appearance feature and the density are defined as follows:
\begin{align} \vspace{-0.6cm}
    A\left(\mathbf{x}_k\right)
    &= \mathcal{B}_1 \!\left( \mathbf{f}_1(x_k, y_k) \odot \mathbf{g}_1(z_k) \right) \nonumber \\
    &+ \mathcal{B}_2 \!\left( \mathbf{f}_2(x_k, z_k) \odot \mathbf{g}_2(y_k) \right) \label{eqn:tensorappearance} \\
    &+ \mathcal{B}_3 \!\left( \mathbf{f}_3(y_k, z_k) \odot \mathbf{g}_3(x_k) \right) \text{,} \nonumber
\end{align}\vspace{-0.5cm}
and
\begin{align}
    \sigma(\mathbf{x}_k)
    &= \mathbf{1}^\top \left( \mathbf{h}_1(x_k, y_k) \odot \mathbf{k}_1(z_k) \right) \nonumber \\
    &+ \mathbf{1}^\top \left( \mathbf{h}_2(x_k, z_k) \odot \mathbf{k}_2(y_k) \right) \label{eqn:tensoropacity} \\
    &+ \mathbf{1}^\top \left( \mathbf{h}_3(y_k, z_k) \odot \mathbf{k}_3(x_k) \right) \text{,} \nonumber
\end{align}
\textit{where} $\sigma(\mathbf{x}_k)$ is the predicted sigma for point $\mathbf{x}_k$, $A\left(\mathbf{x}_k\right)$ is the predicted appearance feature. As shown in VM decomposition, $\mathbf{f}_j$ and $\mathbf{h}_j$ are vector-valued matrix with output dimension $M$ indexed by spatial coordinates, and $\mathbf{g}_j$ and $\mathbf{k}_j$ are vector-valued functions with output dimension $M$ indexed by spatial coordinates. The final RGB color $\mathbf{c}$ is regressed from a tiny MLP with appearance feature and view direction as inputs. With points' density $\sigma$ and colors $\mathbf{c}$, images can be rendered via our composite volumetric rendering equation. We refer readers to TensoRF \cite{Chen2022b} for more details.

For each object scale-invariant representation, the resolution of the volumetric radiance field varies from $160^3$ to $640^3$, regarding the number of objects in the whole scene and GPU memory limit. We use 16 and 48 components to represent density and appearance features respectively. Following TensoRF, we optimize this representation in a coarse-to-fine manner, by starting with a ${0.4}^3 \times$ smaller volume resolution and progressively upsampling the resolution along the training.

\subsection{Object Scale Network}
The network is defined as follows:
\begin{equation}
     p = f_{mlp}(s_1,...,s_K)
\end{equation}
\textit{where} the MLP is implemented as 4 hidden layers with 64 nodes. The scale of the $1^{st}$ object (the static background is chosen in our implementation) is fixed and therefore abandoned from the network input. The last MLP layer is followed by a Sigmoid function.

\begin{table*}[t]
\caption{The Mean Squared Error (MSE) between the object scales estimated by all methods and the ground truth on Dynamic Indoor Scene dataset (lower is better). The methods are trained with different depth supervision: 1) w/o depth, 2) w/ MiDaS depth, and 3) w/ per-object SfM depth.}
\tabcolsep= 0.2cm 
\centering
\begin{scriptsize}
\begin{tabular}{cl|cccc|c}
    \hline
    Depth Sup. & Method & Gnome House & Dining Table & Chessboard & Factory & Average \\
    \hline
    \multirow{4}{*}{\makecell[c]{1)}} & NSFF\cite{Li2021c} & 0.113 & 0.359 & 0.467 & 0.129 & 0.267 \\
    & DynNeRF\cite{Gao2021} & 0.281 & 0.309 & 0.678 & 0.217 & 0.371 \\
    & TiNeuVox\cite{Fang2022} & 0.308 & 0.364 & 0.406 & 0.203 & 0.320 \\
    & HexPlane\cite{Cao2023a} & 1.156 & 0.361 & 0.264 & 0.653 & 0.609 \\
    \hline
    \multirow{4}{*}{\makecell[c]{2)}} & NSFF\cite{Li2021c} & \underline{0.088} & 0.333 & 0.427 & 0.117 & 0.241 \\
    & DynNeRF\cite{Gao2021} & 0.110 & 0.284 & 0.330 & 0.155 & 0.220 \\
    & TiNeuVox\cite{Fang2022} & 0.211 & 0.116 & 0.173 & 0.152 & 0.163 \\
    & HexPlane\cite{Cao2023a} & 3.871 & 0.933 & 1.797 & 1.685 & 2.072 \\
    \hline
    \multirow{5}{*}{\makecell[c]{3)}} & NSFF\cite{Li2021c} & 0.536 & 0.416 & 0.491 & 0.178 & 0.405 \\
    & DynNeRF\cite{Gao2021} & 0.188 & 0.348 & 0.663 & 0.188 & 0.347 \\
    & TiNeuVox\cite{Fang2022} & 0.224 & 0.358 & 0.761 & 0.378 & 0.430 \\
    & HexPlane\cite{Cao2023a} & 0.456 & 0.326 & 0.464 & 0.647 & 0.473 \\
    & \textbf{\nickname{}(Ours)} & \textbf{0.026} & \underline{0.152} & \textbf{0.043} & \textbf{0.036} & \textbf{0.064} \\
    \hline
    2)+3) & Total-Recon\cite{Song2023} & 0.105 & \textbf{0.121} & \underline{0.113} & \underline{0.115} & \underline{0.114} \\
    \hline
\end{tabular}\vspace{-0.3cm}
\end{scriptsize}
\label{tab:scale_err}
\end{table*}

\section{Details of the Joint Optimization}

\textbf{Loss Weights:} In Stage 1 - Boostrapping Per-object Representation, for each object, the RGB loss $\ell_{rgb}^k$ and depth loss $\ell_{depth}^k$ are weighted by $\{1.0, 1.0\}$. In Stage 2 - Alternative Optimization, the RGB loss $\ell_{rgb}^{scene}$, the depth loss $\ell_{depth}^{scene}$, and the segmentation loss $\ell_{seg}^{scene}$ are weighted by $\{1.0, 1.0, 0.01\}$ in the whole training process.

\textbf{Training Schedule:} We adopt the Adam optimizer with a learning rate of 0.001 for both object scale-invariant representation module and the object scale network. We optimize the former for a total of 30K/ 30K/ 80K/ iterations on the Dynamic Indoor Scene/ Oxford Multimotion/ Nvidia Dynamic Scene datasets. To train on each scene, we take the first 1K iterations as the Stage 1 (Boostrapping). Then, we take 1K iterations for optimizing object scale-invariant representation networks, followed by 1K iterations for training the object scale network.

\begin{figure*}[t]
\centering
  \includegraphics[width=2.0\columnwidth]{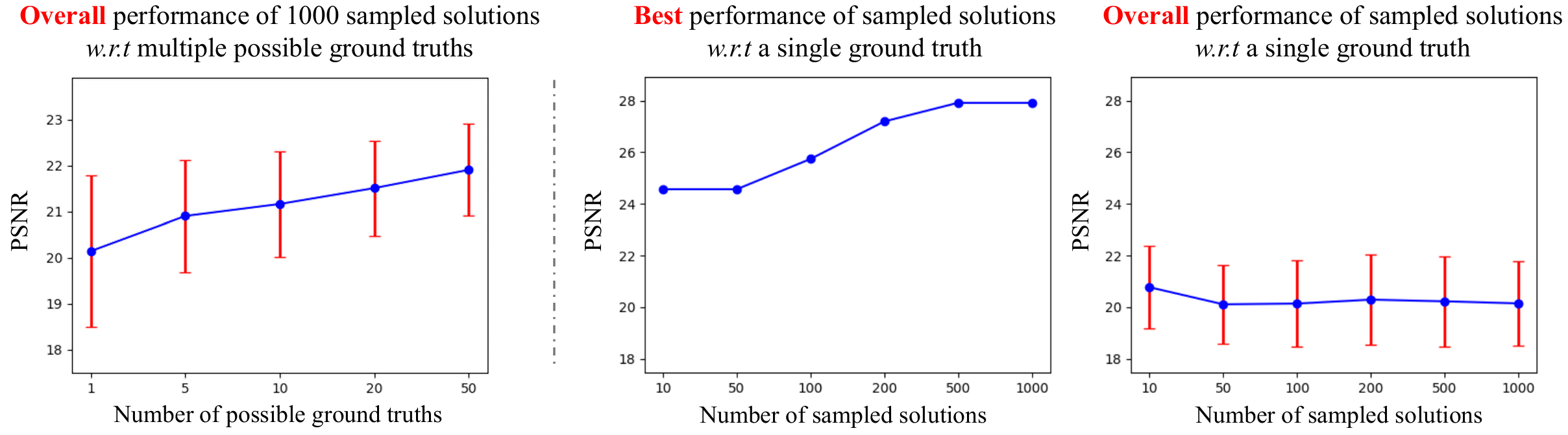}
  \caption{\textbf{Left}: Overall performance of 1000 sampled solutions \wrt{} different number of possible ground truths. \textbf{Right}: Best and overall performance of different number of solutions \wrt{} a single ground truth.}
  \label{fig:stats}
\end{figure*}

\begin{table*}[h]
\caption{Quantitative results of our \nickname{} with different object orders on the ``Gnome House" scene.}
\tabcolsep= 0.2cm 
\centering
\begin{scriptsize}
\begin{tabular}{c|cccccc}
    \hline
    permutations & PSNR$\uparrow$ & SSIM$\uparrow$ & LPIPS$\downarrow$ & SSIMAE$\downarrow$ & PQ$\uparrow$ & mIoU$\uparrow$ \\
    \hline
    \{1, 2, 3\} & 27.917 & 0.833 & 0.106 & 0.079 & 95.889 & 95.889 \\
    \{1, 3, 2\} & 28.045 & 0.834 & 0.108 & 0.081 & 95.856 & 95.856 \\
    \{2, 1, 3\} & 27.954 & 0.834 & 0.108 & 0.081 & 95.896 & 95.896 \\
    \{2, 3, 1\} & 27.899 & 0.833 & 0.107 & 0.078 & 95.863 & 95.863 \\
    \{3, 1, 2\} & 27.976 & 0.834 & 0.109 & 0.085 & 95.827 & 95.827 \\
    \{3, 2, 1\} & 27.893 & 0.833 & 0.107 & 0.081 & 95.869 & 95.869 \\
    \hline
    Average & 27.947$\pm$0.058 & 0.834$\pm$0.001 & 0.108$\pm$0.001 & 0.081$\pm$0.002 & 95.867$\pm$0.025 & 95.867$\pm$0.025 \\ 
    \hline
\end{tabular}\vspace{-0.3cm}
\end{scriptsize}
\label{tab:exp_perm}
\end{table*}

\section{Details of Datasets}

\textbf{Dynamic Indoor Scene Dataset:} We directly adopt the 4 scenes used by NVFi \cite{Li2023c}: ``Gnome House", ``Dining Table", ``Chessboard" and ``Factory". Since some objects in the original dataset are hardly visible in a monocular video, we re-compose the objects and their motions after requesting the original 3D mesh models from the authors. For each scene, we collect RGB images at 15 viewpoints over 1 second, where each viewpoint has 15 frames captured. We simulate a moving monocular camera by extracting frames from each viewpoint at different time instances, \ie, 15 frames for the training split, while leaving the 210 frames at held-out viewpoints and time instances for the testing split. 

\textbf{Oxford Multimotion Dataset:} Among several sequences in the dataset, we select 4 scenes: ``swinging\_4\_translational", ``swinging\_4\_unconstrained", ``occlusion\_2\_translational", and ``occlusion\_2\_unconstrained". The first two scenes have 4 cubes hanged from the ceiling, swinging and flipping along different directions. The last two scenes have a large block tower and a small block moving left to right, with the latter occasionally being occluded by the former. For each scene, we extract a 15-frame clip over 2 seconds, taking 15 frames from the left camera for training and leaving 15 frames from the right camera for testing.

\textbf{NVIDIA Dynamic Scene Dataset:} Among 8 scenes in the original dataset, we select 3 scenes ``Balloon2", ``Skating", and ``Truck", where motions of balloon, skateboarder, and truck are all rigid or approximately rigid. Each scene is captured by 12 synchronized cameras, and we follow the evaluation protocol in DynamicNeRF \cite{Gao2021} to train on 12 frames sampled from each camera viewpoint at different time instances, and test on 12 frames sampled from the first camera viewpoint at different time instances.

\section{Evaluation of Scale Estimation}

For a more in-depth analysis of our superior performance, we compare the object scales estimated by different methods with the ground truth on the synthetic Dynamic Indoor Scene dataset, measured by Mean Squared Error (MSE). For baselines modeling deformable objects, the estimated scales are computed from the corresponding object regions in rendered depths. As shown in Table \ref{tab:scale_err}, our object scale network is crucial to recover realistic 3D scenes, as demonstrated by the superior estimated object scales over baselines.

\section{Evaluation of Invariance to Object Orders}

Our whole framework is designed in a scene-specific optimization fashion akin to NeRF. Before training our OSN, the input scales $\{s_1 \cdots s_k \cdots s_K\}$ are flexible to be associated with any objects thanks to the inherent symmetry of MLP architecture. The subsequent rendering algorithms are also invariant to permutations of objects. This means that our framework does not rely on a specific object order.

We further evaluate such ability of our framework on the ``Gnome House" scene. Particularly, we separately train 6 scale networks on 6 different permutations of the 3 objects. We then sample $100K$ scale combinations and feed them into the trained 6 scale networks, obtaining 6 averaged scores for all permutations. Such 6 scores show a standard deviation of 0.0042 over the 6 permutations on the overall $100K$ samples, which is mainly caused by random initialization of 6 networks, and therefore negligible. This validates that our scale network does not rely on a specific order of objects.

We further feed a specific scale combination to the trained object scale-invariant representations for rendering. Table \ref{tab:exp_perm} shows that all 6 permutations have nearly the same results, validating that the object scale-invariant representations and rendering algorithms doesn't rely on the object order as well.

\section{Analysis on Number of Possible Ground Truths and Sampled Solutions}

We evaluate the overall performance of all 1000 samples \wrt{} different number of possible ground truths on ``Gnome House" scene of the Dynamic Indoor Scene Dataset. As shown in the error bar statistics in Figure \ref{fig:stats} (left), when there is only a single ground truth, the performance of 1000 samples shows large variance since most of them represent totally different scenes. When the number of ground truths increases to 5, 10, 20, and 50, more and more samples can match proper ground truth 3D scenes.

We also evaluate the best and overall (averaged) performance of sampled solutions given various number of samples, \wrt{} a single ground truth. As shown in Figure \ref{fig:stats} (right), we can see that: a) For the best performance, the PSNR score gradually increases since a singular ground truth is more likely to be better approached by more samples. b) For the overall (averaged) performance, it is actually not meaningful \wrt{} to a single ground truth, since sampled solutions are free to represent any possible meaningful 3D scene behind the same monocular video. In the plot, both its average and standard deviation show little changes given various number of samples, because the majority of these solutions could deviate from the singular ground truth.

\begin{table*}[h] \vspace{-0.4cm}
\caption{Quantitative results for novel view synthesis on the ``Chessboard++" scene. The methods are trained with different depth supervision: 1) w/o depth, 2) w/ MiDaS depth, and 3) w/ per-object SfM depth.}
\centering
\begin{scriptsize}
\begin{tabular}{cl|cccc}
    \hline
    \multirow{2}{*}{} & & \multicolumn{4}{c}{Chessboard++} \\
    Depth Sup. & Method & PSNR$\uparrow$ & SSIM$\uparrow$ & LPIPS$\downarrow$ & SSIMAE$\downarrow$ \\
    \hline
    \multirow{4}{*}{\makecell[c]{1)}} & NSFF\cite{Li2021c} & 18.918 & 0.671 & 0.365 & 0.921  \\
    & DynNeRF\cite{Gao2021} & 20.521 & 0.752 & 0.290 & 0.595  \\
    & TiNeuVox\cite{Fang2022} & 20.135 & 0.627 & 0.452 & 0.703  \\
    & HexPlane\cite{Cao2023a} & 16.862 & 0.524 & 0.628 & 2.056  \\
    \hline
    \multirow{4}{*}{\makecell[c]{2)}} & NSFF\cite{Li2021c} & 18.766 & 0.620 & 0.414 & 0.805  \\
    & DynNeRF\cite{Gao2021} & 20.292 & 0.742 & 0.299 & 0.613  \\
    & TiNeuVox\cite{Fang2022} & 20.159 & 0.625 & 0.468 & 0.589  \\
    & HexPlane\cite{Cao2023a} & 16.919 & 0.547 & 0.424 & 1.159  \\
    \hline
    \multirow{5}{*}{\makecell[c]{3)}} & NSFF\cite{Li2021c} & 19.581 & 0.733 & 0.308 & 0.733  \\
    & DynNeRF\cite{Gao2021} & 20.773 & 0.757 & 0.284 & 0.538  \\
    & TiNeuVox\cite{Fang2022} & 20.233 & 0.645 & 0.445 & 0.738  \\
    & HexPlane\cite{Cao2023a} & 19.998 & 0.618 & 0.446 & 0.562  \\
    & \textbf{\nickname{}(Ours)} & \textbf{21.900} & \textbf{0.807} & \textbf{0.241} & \textbf{0.337} \\
    \hline
    2)+3) & Total-Recon\cite{Song2023} & 20.908 & 0.765 & 0.253 & 0.444  \\
    \hline
\end{tabular}
\end{scriptsize}
\label{tab:nvs_chesspp} \vspace{-0.2cm}
\end{table*}

\begin{table*}[h]
\caption{Quantitative results of all methods on real-world Sequence 0007 of KITTI dataset. The methods are trained with different depth supervision: 1) w/o depth, 2) w/ MiDaS depth, and 3) w/ per-object SfM depth.}
\tabcolsep= 0.2cm 
\centering
\begin{scriptsize}
\begin{tabular}{cl|cccc|ccc}
    \hline
    \multirow{2}{*}{} & &  \multicolumn{3}{c}{KITTI - Sequence 0007} \\
    Depth Sup. & Method & SNR$\uparrow$ & SSIM$\uparrow$ & LPIPS$\downarrow$ \\
    \hline
    \multirow{4}{*}{\makecell[c]{1)}} & NSFF\cite{Li2021c} & 16.668 & 0.612 & 0.349 \\
    & DynNeRF\cite{Gao2021} & 17.437 & 0.635 & 0.376 \\
    & TiNeuVox\cite{Fang2022} & 14.874 & 0.536 & 0.431 \\
    & HexPlane\cite{Cao2023a} & 15.935 & 0.597 & 0.435 \\
    \hline
    \multirow{4}{*}{\makecell[c]{2)}} & NSFF\cite{Li2021c} & 16.649 & 0.607 & 0.359 \\
    & DynNeRF\cite{Gao2021} & 17.121 & 0.611 & 0.405 \\
    & TiNeuVox\cite{Fang2022} & 16.539 & 0.580 & 0.401 \\
    & HexPlane\cite{Cao2023a} & 15.189 & 0.551 & 0.472 \\
    \hline
    \multirow{5}{*}{\makecell[c]{3)}} & NSFF\cite{Li2021c} & 17.602 & 0.648 & 0.360 \\
    & DynNeRF\cite{Gao2021} & 16.955 & 0.620 & 0.390 \\
    & TiNeuVox\cite{Fang2022} & 16.290 & 0.580 & 0.399 \\
    & HexPlane\cite{Cao2023a} & 16.738 & 0.635 & 0.394 \\
    & \textbf{\nickname{}(Ours)}  & \underline{18.492} & \underline{0.684} & \underline{0.262} \\
    \hline
    2)+3) & Total-Recon\cite{Song2023} & \textbf{18.508} & \textbf{0.689} & \textbf{0.258} \\
    \hline
\end{tabular}\vspace{-0.3cm}
\end{scriptsize}
\label{tab:nvs_kitti}
\end{table*}

\section{Evaluation on Large and Variable Number of Objects}

\begin{figure*}[t]
\centering
  \includegraphics[width=2.0\columnwidth]{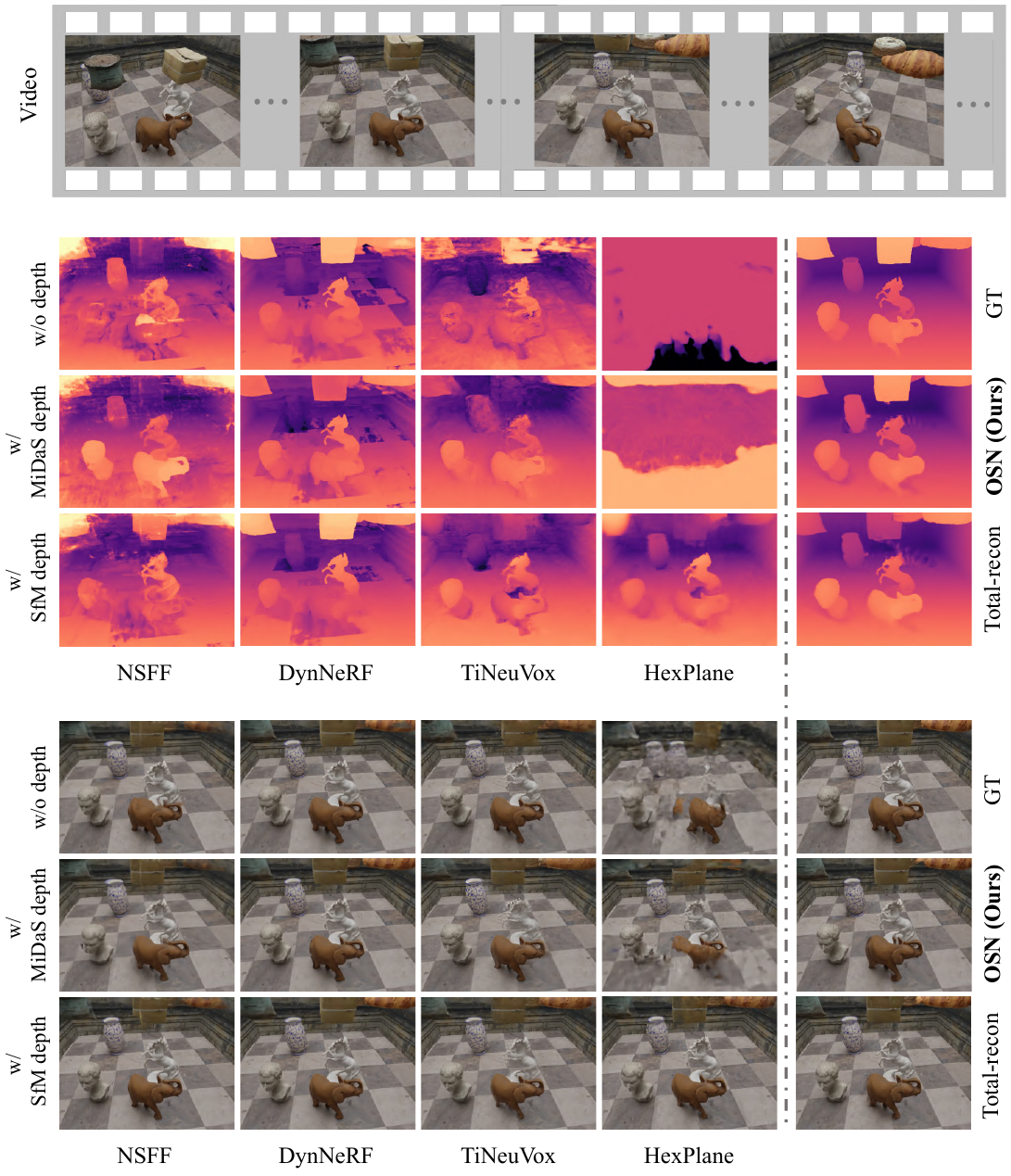}
  \caption{Qualitative results of dynamic novel view RGB/depth synthesis on the synthetic ``Chessboard++" scene.}
  \label{fig:qual_res_chesspp}
\end{figure*}

By design, our object scale network is not restricted to the number of dynamic objects. Given more objects, we may just need more MLP layers to enhance its ability. We further evaluate our current object scale network (4 MLP layers) on more objects. Particularly, we build another synthetic scene ``Chessboard++" containing 8 moving objects (twice the maximum number of moving objects in previous scenes). During capturing the video, some objects get in or out of the view, resulting in an extremely challenging monocular video with large and dynamic (variable) number of objects. Table \ref{tab:nvs_chesspp} shows quantitative results of all methods. Our method achieves the best performance, demonstrating the capability of tackling large and variable number of objects. Qualitative results are in Figure \ref{fig:qual_res_chesspp}.

\section{Evaluation on Self-Driving Scenes}

\begin{figure*}[t]
\centering
  \includegraphics[width=2.0\columnwidth]{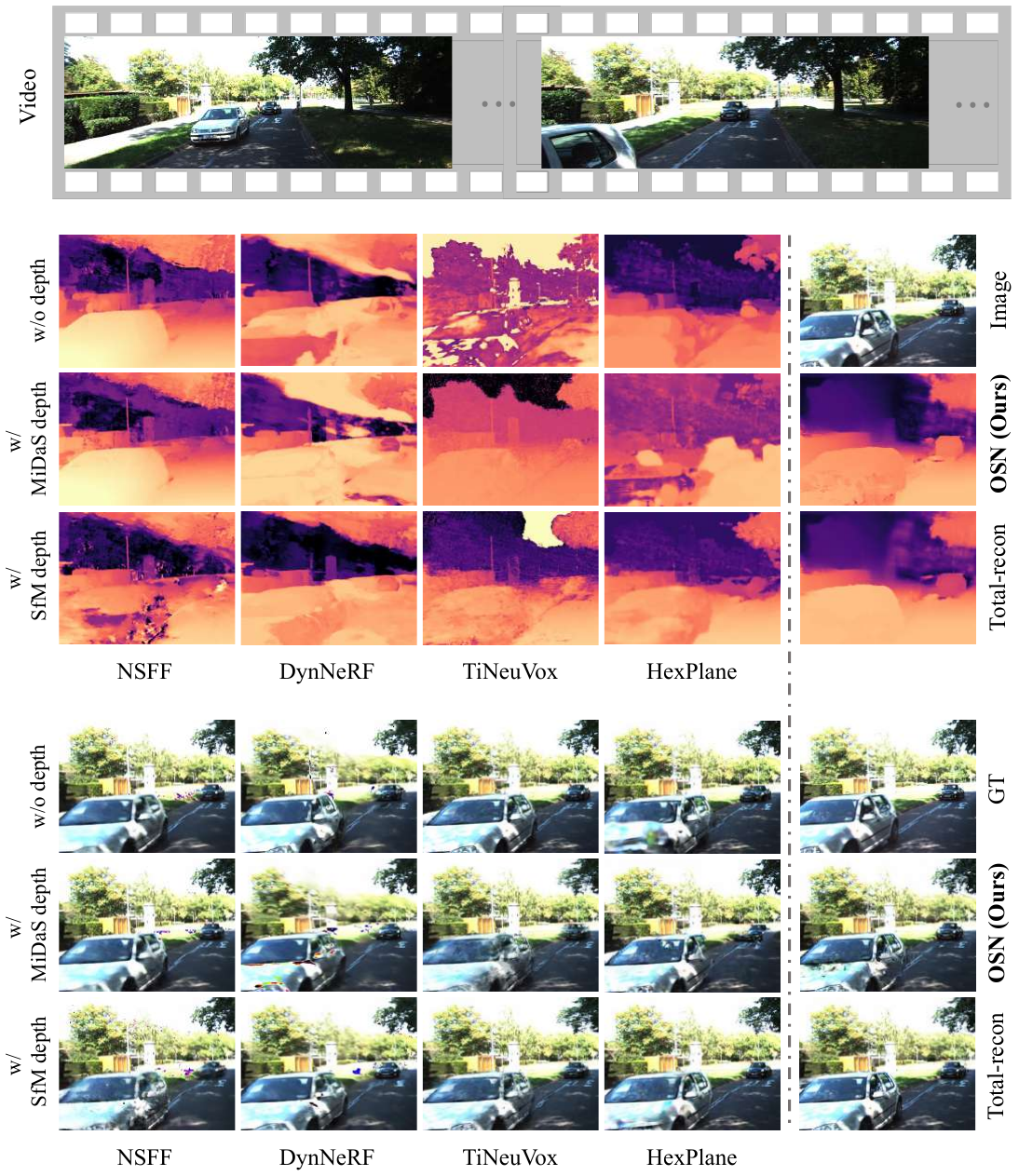}
  \caption{Qualitative results of dynamic novel view RGB/depth synthesis on a selected scene from Sequence 0007 of KITTI dataset.}
  \label{fig:qual_res_kitti}
\end{figure*}

We extract a 15-frame clip from the Sequence 0007 of KITTI dataset for evaluation. As shown in Table \ref{tab:nvs_kitti} and Figure \ref{fig:qual_res_kitti}, our method can tackle the challenging real outdoor scenes and achieves competing performance. Nevertheless, our method is not primarily designed for autonomous driving scenarios as we have discussed in Section \ref{sec:app_scene}. In fact, while examining scenes in KITTI dataset, we find that the common forward-moving camera motions can easily result in failures in SfM or object tracking, making the preliminarily processed data (Section \ref{sec:method_preliminary}) less satisfactory for our method.

\section{Evaluation on Daily Scenes}

\begin{figure*}[t]
\centering
  \includegraphics[width=2.0\columnwidth]{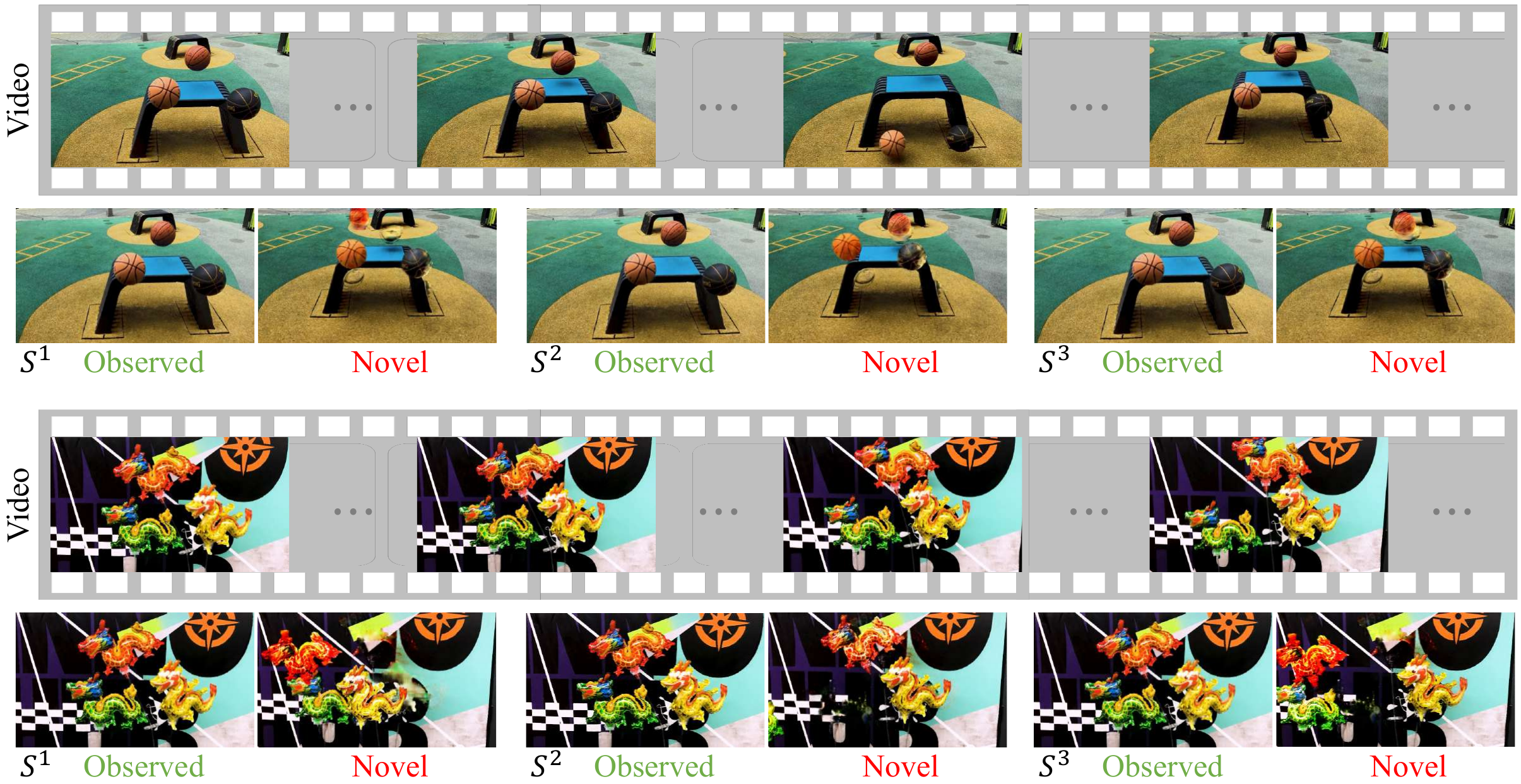}
  \caption{Qualitative results on two daily scenes captured by a mobile phone. For each scene we show the captured monocular video, and the rendering results from observed and novel views under 3 sampled scale combinations at a specific time step.}
  \label{fig:qual_res_iphone}
\end{figure*}

We additionally capture two daily scenes, ``Bouncing Basketballs" and ``Flying Dragon-Balloons", using the monocular camera on a mobile phone and show qualitative results in Figure \ref{fig:qual_res_iphone}, demonstrating real-world applications of our method.

\clearpage
\onecolumn
\section{More Quantitative and Qualitative Results}
\begin{figure*}[t]
    \centering    \includegraphics[width=1.0\columnwidth]{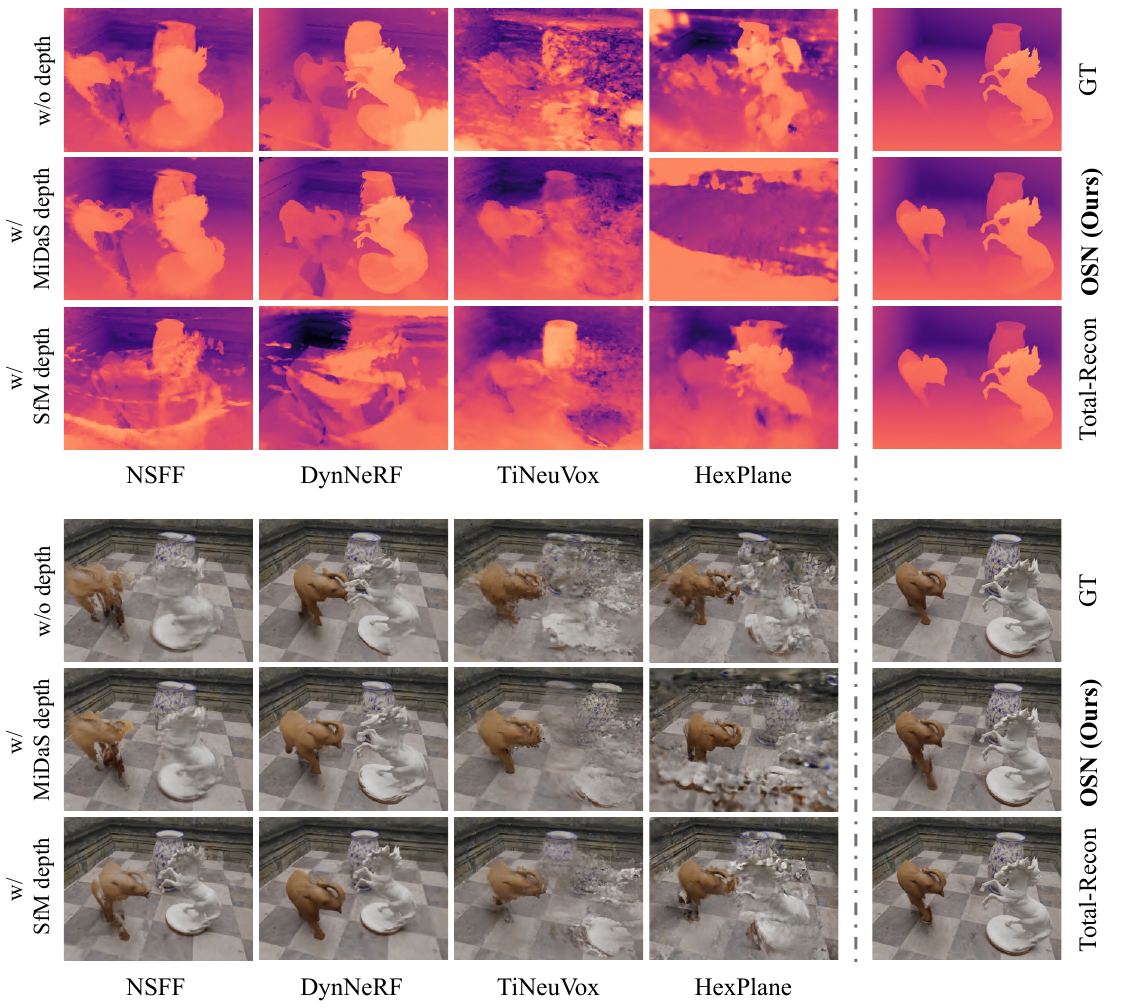}
    \vskip -0.18in
    \caption{Qualitative results of dynamic novel view RGB/depth synthesis on the ``Chessboard" of Dynamic Indoor Scene Dataset.}
    \label{fig:chessboard}
    \vskip -0.2in
\end{figure*}

\begin{figure*}[t]
    \centering    \includegraphics[width=1.0\columnwidth]{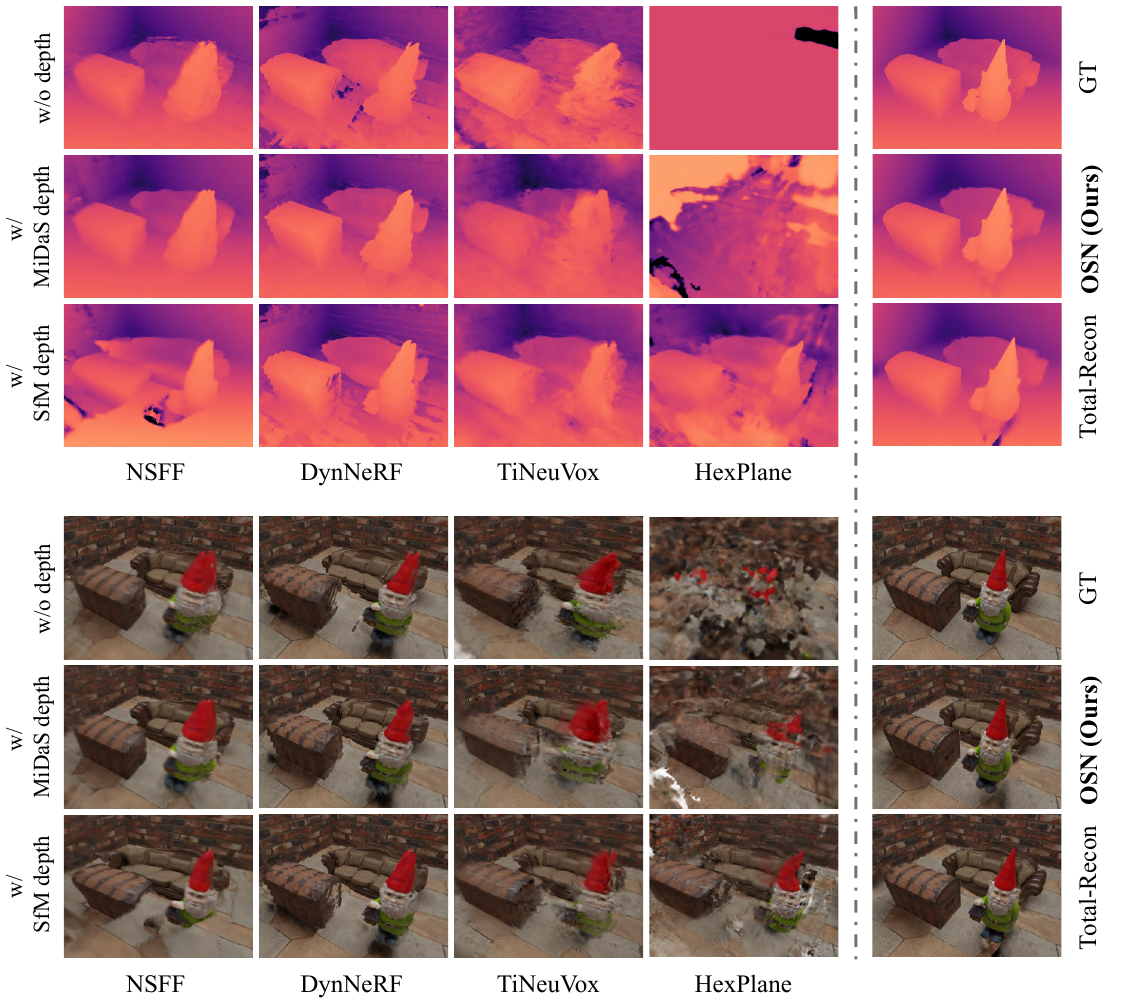}
    \vskip -0.18in
    \caption{Qualitative results of dynamic novel view RGB/depth synthesis on the ``Gnome House" of Dynamic Indoor Scene Dataset.}
    \label{fig:darkroom}
    \vskip -0.2in
\end{figure*}

\begin{figure*}[t]
    \centering    \includegraphics[width=1.0\columnwidth]{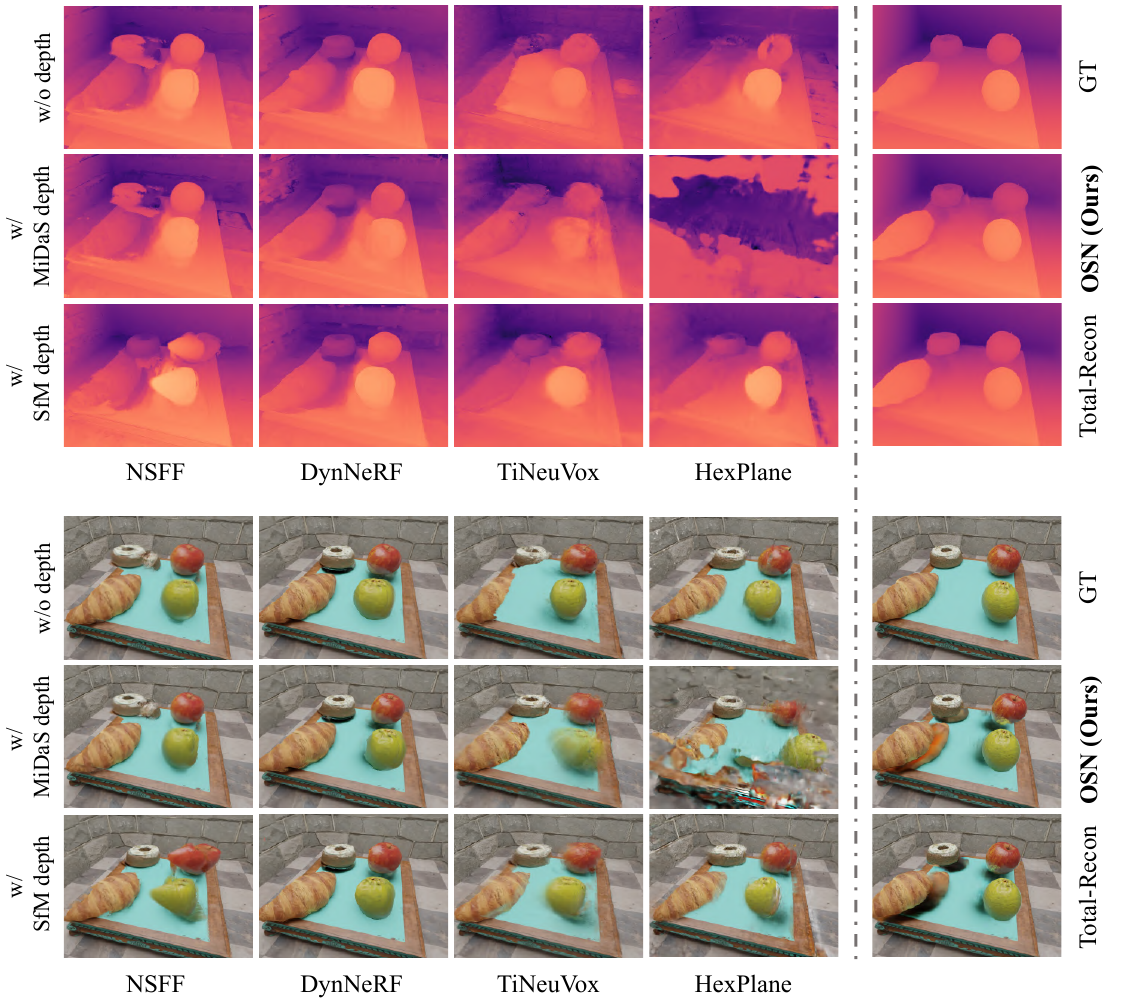}
    \vskip -0.18in
    \caption{Qualitative results of dynamic novel view RGB/depth synthesis on the ``Dining Table" of Dynamic Indoor Scene Dataset.}
    \label{fig:dining}
    \vskip -0.2in
\end{figure*}

\begin{figure*}[t]
    \centering    \includegraphics[width=1.0\columnwidth]{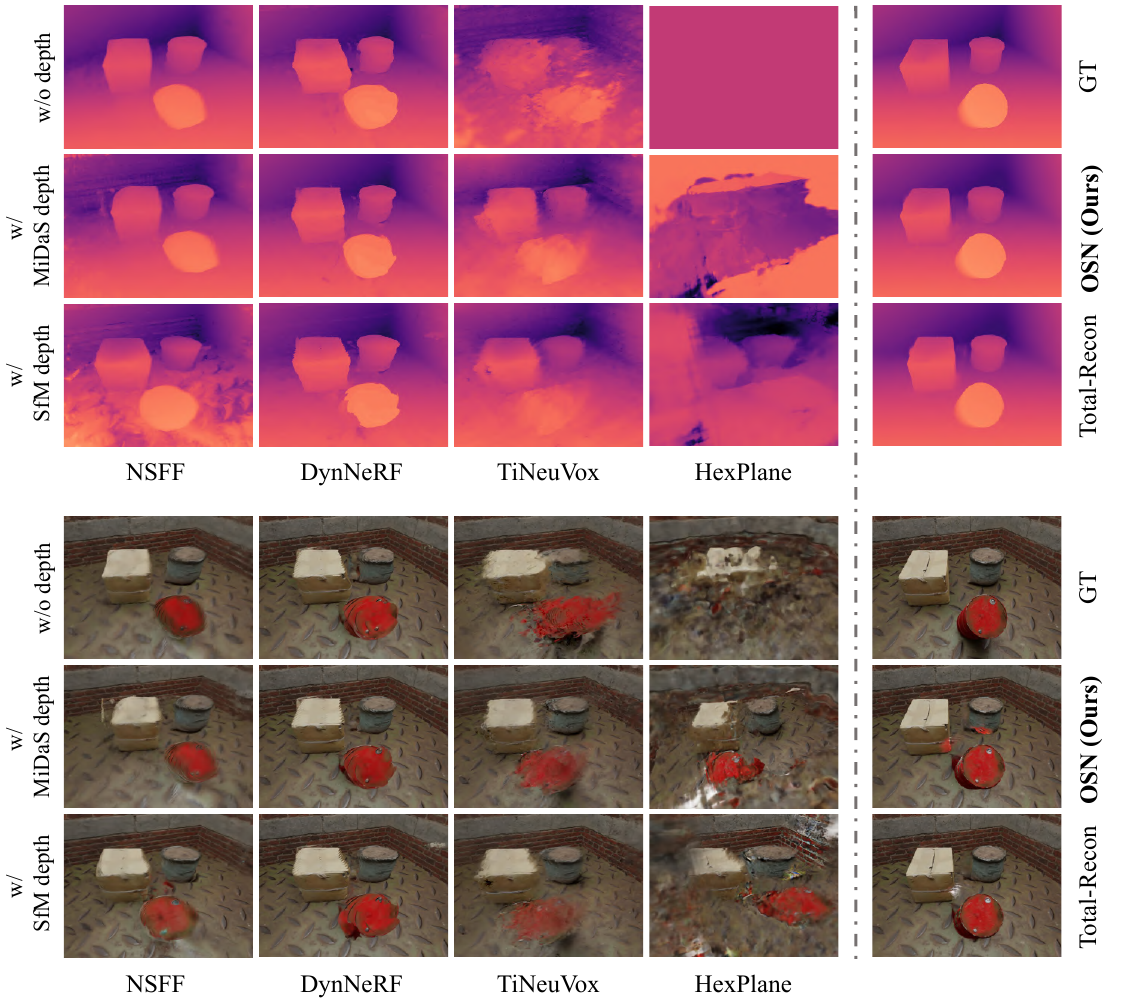}
    \vskip -0.18in
    \caption{Qualitative results of dynamic novel view RGB/depth synthesis on the ``Factory" of Dynamic Indoor Scene Dataset.}
    \label{fig:factory}
    \vskip -0.2in
\end{figure*}

\begin{figure*}[t]
    \centering    \includegraphics[width=1.0\columnwidth]{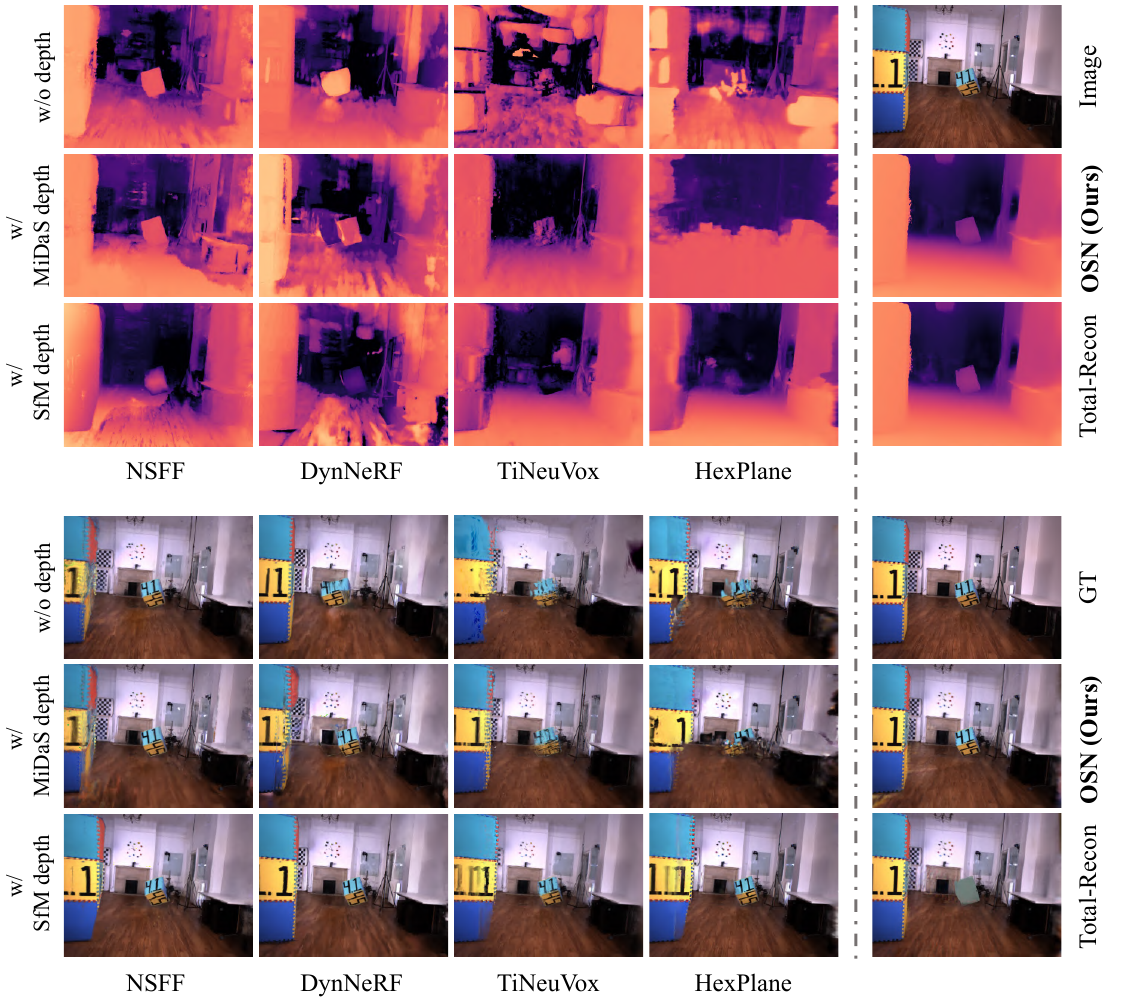}
    \vskip -0.18in
    \caption{Qualitative results of dynamic novel view RGB/depth synthesis on the ``occlusion\_2\_translational" of Oxford Multimotion Dataset.}
    \label{fig:occlusionT}
    \vskip -0.2in
\end{figure*}

\begin{figure*}[t]
    \centering    \includegraphics[width=1.0\columnwidth]{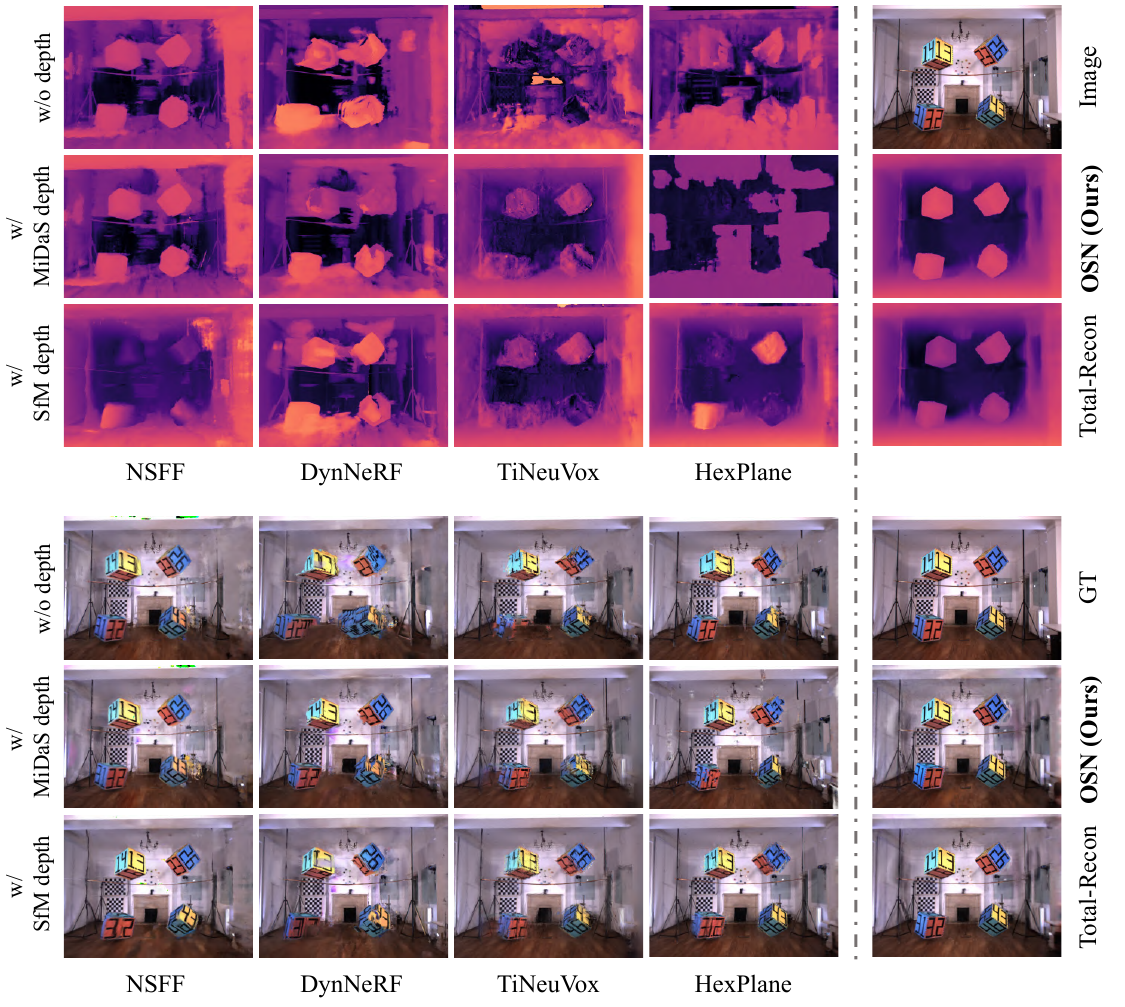}
    \vskip -0.18in
    \caption{Qualitative results of dynamic novel view RGB/depth synthesis on the ``swinging\_4\_translational" of Oxford Multimotion Dataset.}
    \label{fig:swingingT}
    \vskip -0.2in
\end{figure*}

\begin{figure*}[t]
    \centering    \includegraphics[width=1.0\columnwidth]{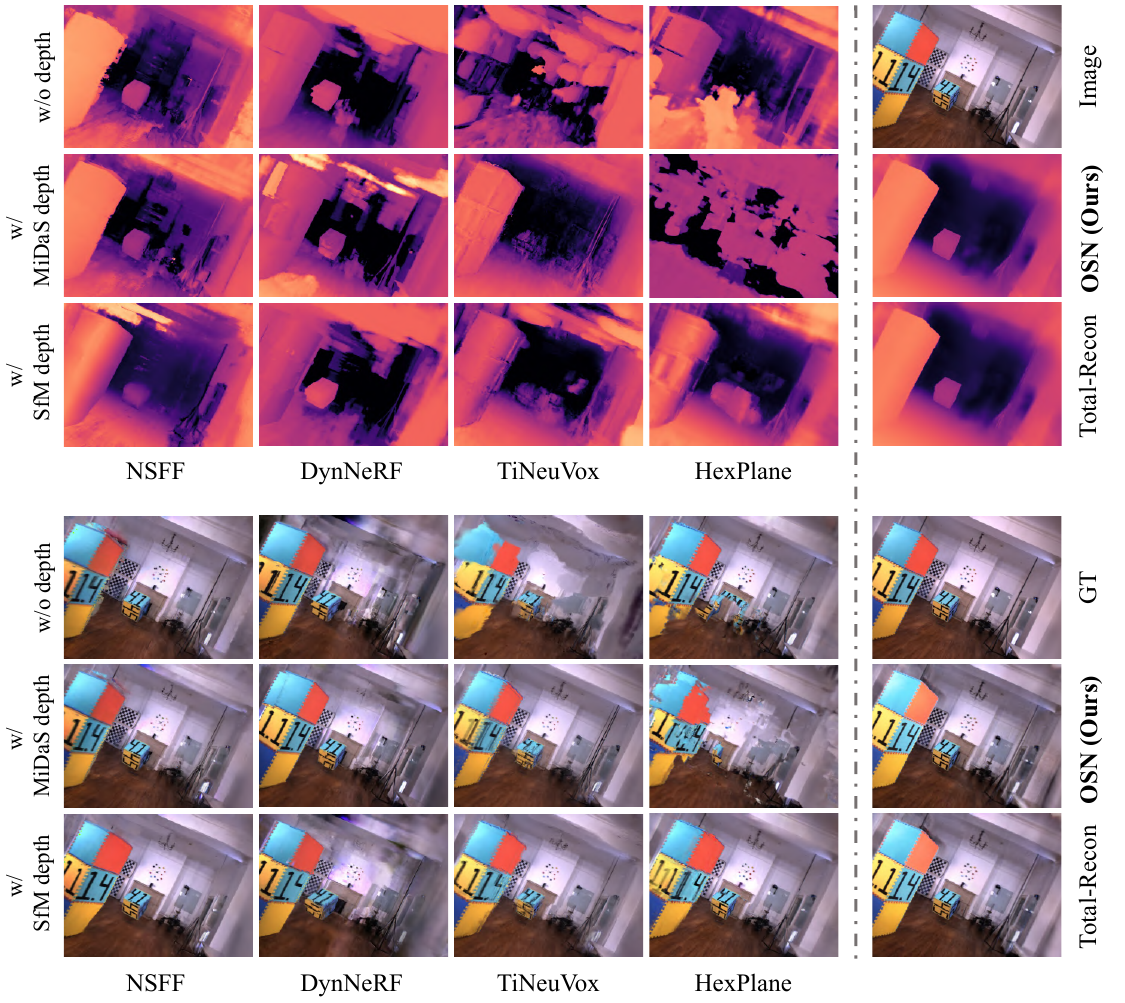}
    \vskip -0.18in
    \caption{Qualitative results of dynamic novel view RGB/depth synthesis on the ``occlusion\_2\_unconstrained" of Oxford Multimotion Dataset.}
    \label{fig:occlusionU}
    \vskip -0.2in
\end{figure*}

\begin{figure*}[t]
    \centering    \includegraphics[width=1.0\columnwidth]{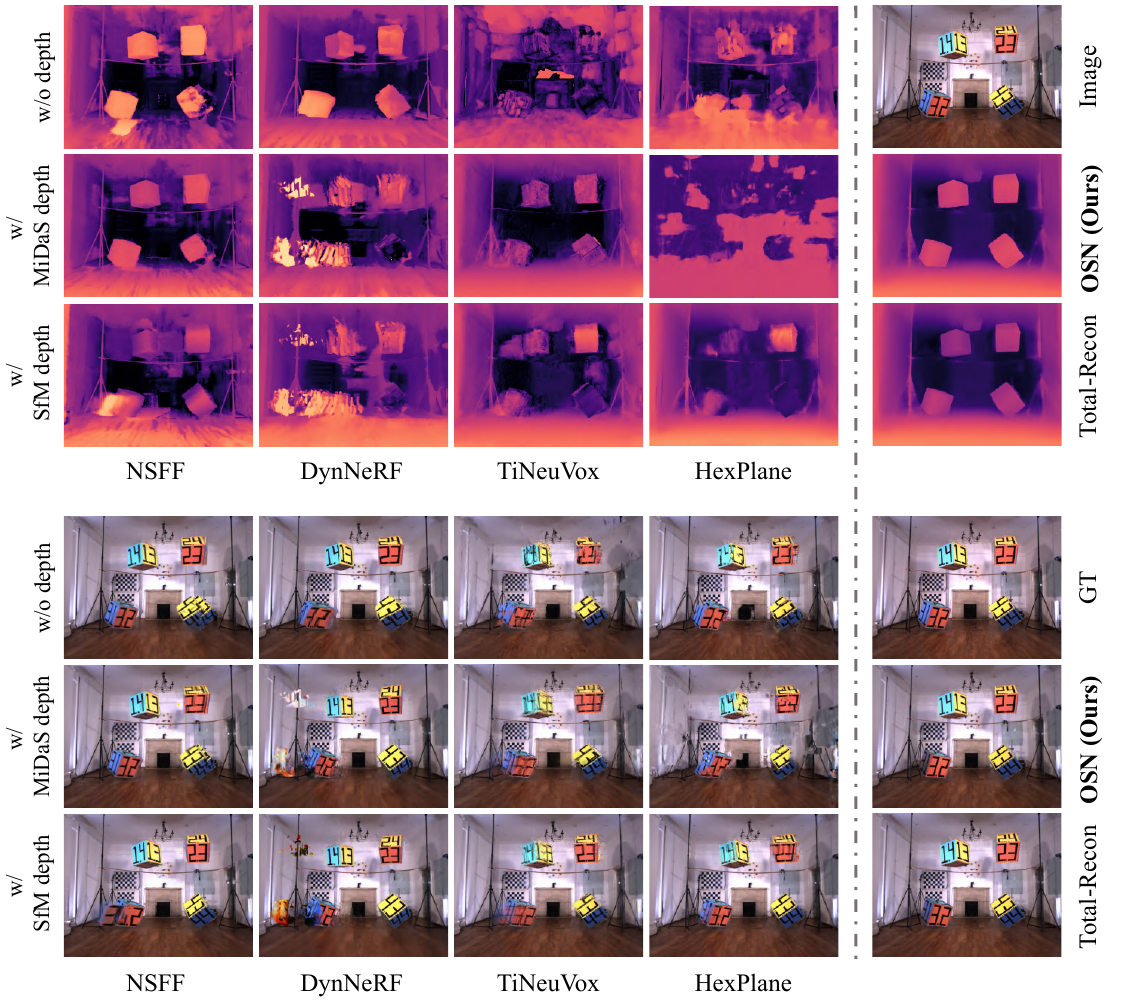}
    \vskip -0.18in
    \caption{Qualitative results of dynamic novel view RGB/depth synthesis on the ``swinging\_4\_unconstrained" of Oxford Multimotion Dataset.}
    \label{fig:swingingU}
    \vskip -0.2in
\end{figure*}

\begin{figure*}[t]
    \centering    \includegraphics[width=1.0\columnwidth]{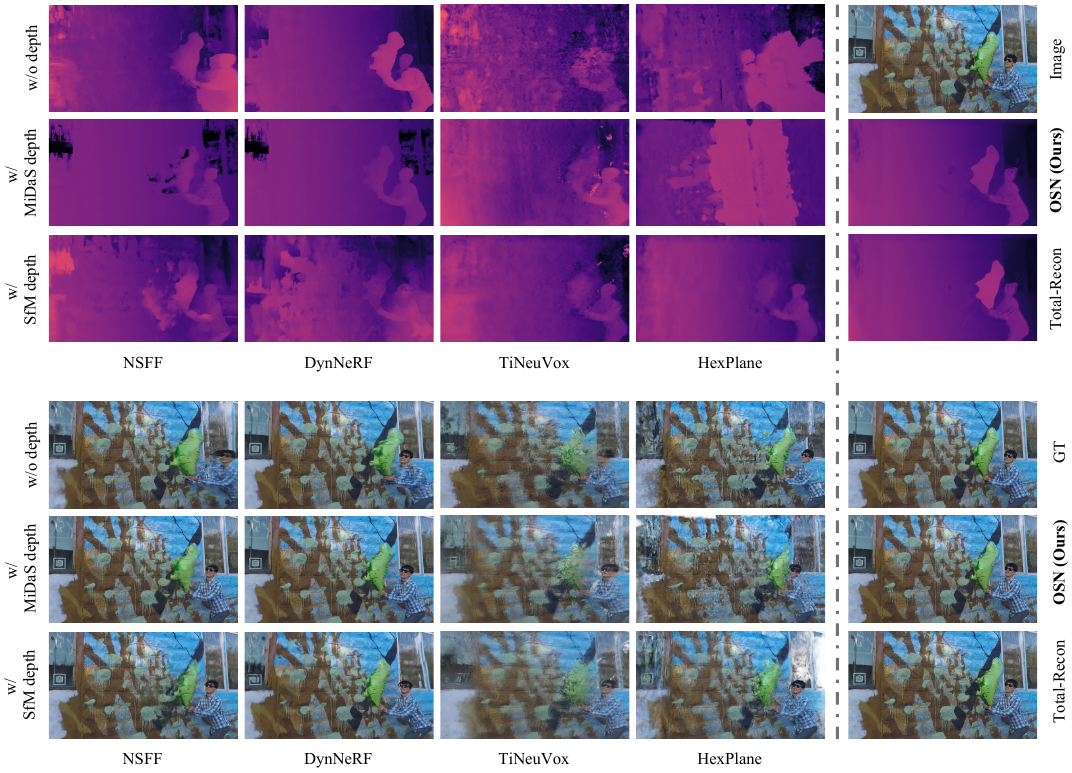}
    \vskip -0.18in
    \caption{Qualitative results of dynamic novel view RGB/depth synthesis on the ``Balloon2" of NVIDIA Dynamic Scene Dataset.}
    \label{fig:Balloon2}
    \vskip -0.2in
\end{figure*}

\begin{figure*}[t]
    \centering    \includegraphics[width=1.0\columnwidth]{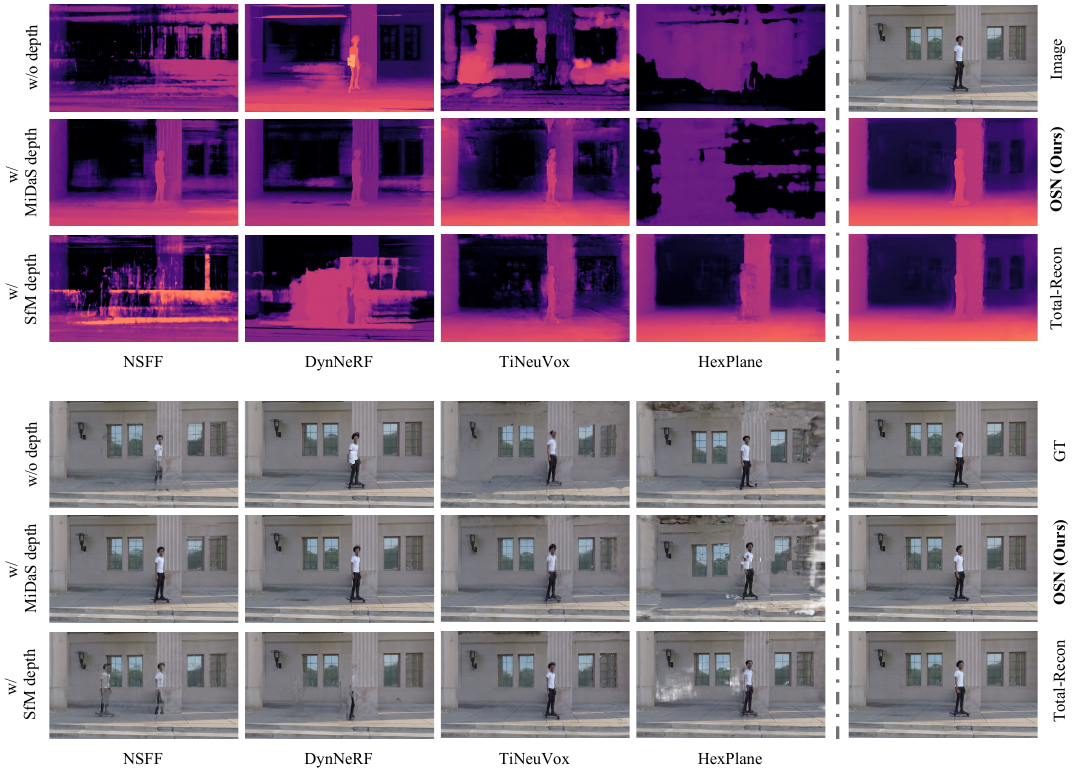}
    \vskip -0.18in
    \caption{Qualitative results of dynamic novel view RGB/depth synthesis on the ``Skating" of NVIDIA Dynamic Scene Dataset.}
    \label{fig:Skating}
    \vskip -0.2in
\end{figure*}

\begin{figure*}[t]
    \centering    \includegraphics[width=1.0\columnwidth]{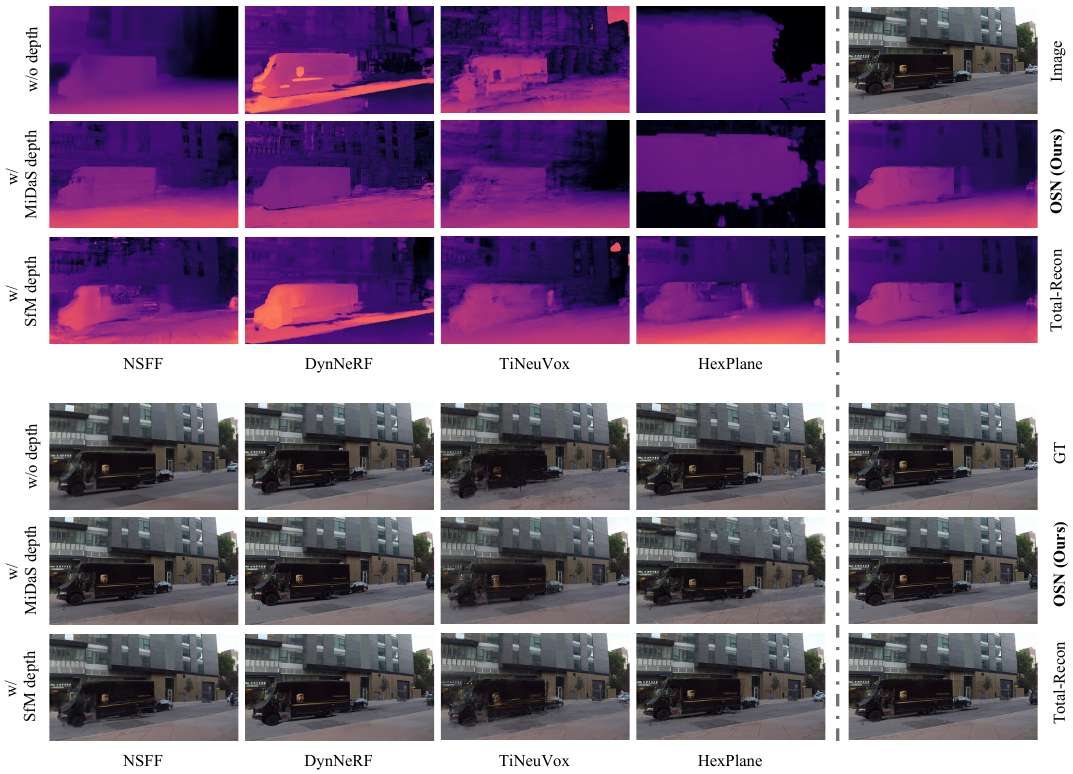}
    \vskip -0.18in
    \caption{Qualitative results of dynamic novel view RGB/depth synthesis on the ``Truck" of NVIDIA Dynamic Scene Dataset.}
    \label{fig:Truck}
    \vskip -0.2in
\end{figure*}

We report the quantitative results of novel view synthesis for individual scenes in the three datasets. As shown in Tables \ref{tab:nvs_indoor}/\ref{tab:nvs_multimotion}/\ref{tab:nvs_dynamicscene} and Figures \ref{fig:chessboard}/\ref{fig:darkroom}/\ref{fig:dining}/\ref{fig:factory}/\ref{fig:occlusionT}/\ref{fig:swingingT}/\ref{fig:occlusionU}/\ref{fig:swingingU}/\ref{fig:Balloon2}/\ref{fig:Skating}/\ref{fig:Truck}, our method achieves excellent performance, especially in learning fine-grained 3D geometry which can be seen from our high-qualty depth estimation. 
 
\begin{table*}[h] \vspace{-0.4cm}
\caption{Per-scene quantitative results for novel view synthesis on the Dynamic Indoor Scene Dataset. The methods are trained with different depth supervision: 1) w/o depth, 2) w/ MiDaS depth, and 3) w/ per-object SfM depth.}
\centering
\begin{scriptsize}
\begin{tabular}{cl|cccc|cccc}
    \hline
    \multirow{2}{*}{} & & \multicolumn{4}{c}{Gnome House} & \multicolumn{4}{c}{Chessboard} \\
    Depth Sup. & Method & PSNR$\uparrow$ & SSIM$\uparrow$ & LPIPS$\downarrow$ & SSIMAE$\downarrow$ & PSNR$\uparrow$ & SSIM$\uparrow$ & LPIPS$\downarrow$ & SSIMAE$\downarrow$ \\
    \hline
    \multirow{4}{*}{\makecell[c]{1)}} & NSFF\cite{Li2021c} & 24.695 & 0.717 & 0.304 & 0.164 & 18.285 & 0.701 & 0.360 & 0.735 \\
    & DynNeRF\cite{Gao2021} & 21.904 & 0.674 & 0.337 & 0.389 & 18.259 & 0.718 & 0.344 & 0.757 \\
    & TiNeuVox\cite{Fang2022} & 22.193 & 0.602 & 0.334 & 0.358 & 19.760 & 0.622 & 0.377 & 0.768 \\
    & HexPlane\cite{Cao2023a} & 16.993 & 0.431 & 0.693 & 1.545 & 18.977 & 0.624 & 0.368 & 0.904 \\
    \hline
    \multirow{4}{*}{\makecell[c]{2)}} & NSFF\cite{Li2021c} & 24.482 & 0.667 & 0.389 & 0.201 & 18.322 & 0.716 & 0.340 & 1.118 \\
    & DynNeRF\cite{Gao2021} & 23.304 & 0.714 & 0.300 & 0.214 & 18.894 & 0.728 & 0.322 & 0.550 \\
    & TiNeuVox\cite{Fang2022} & 23.753 & 0.644 & 0.299 & 0.315 & 20.640 & 0.689 & 0.327 & 0.481 \\
    & HexPlane\cite{Cao2023a} & 19.149 & 0.473 & 0.574 & 1.784 & 17.018 & 0.541 & 0.529 & 1.342 \\
    \hline
    \multirow{5}{*}{\makecell[c]{3)}} & NSFF\cite{Li2021c} & 21.142 & 0.537 & 0.487 & 0.480 & 18.494 & 0.712 & 0.353 & 0.671 \\
    & DynNeRF\cite{Gao2021} & 22.178 & 0.684 & 0.327 & 0.287 & 17.952 & 0.681 & 0.414 & 1.209 \\
    & TiNeuVox\cite{Fang2022} & 22.396 & 0.621 & 0.320 & 0.277 & 19.608 & 0.666 & 0.344 & 0.703 \\
    & HexPlane\cite{Cao2023a} & 20.344 & 0.528 & 0.430 & 0.436 & 19.279 & 0.642 & 0.358 & 0.612 \\
    & \textbf{\nickname{}(Ours)} & \textbf{27.917} & \textbf{0.833} & \textbf{0.106} & \textbf{0.079} & 25.205 & 0.872 & 0.120 & \textbf{0.087} \\
    \hline
    2)+3) & Total-Recon\cite{Song2023} & 25.077 & 0.802 & 0.142 & 0.181 & \textbf{25.606} & \textbf{0.883} & \textbf{0.111} & 0.133 \\
    \hline
    
    \multirow{2}{*}{} & & \multicolumn{4}{c}{Dining Table} & \multicolumn{4}{c}{Factory} \\
    Depth Sup. & Method & PSNR$\uparrow$ & SSIM$\uparrow$ & LPIPS$\downarrow$ & SSIMAE$\downarrow$ & PSNR$\uparrow$ & SSIM$\uparrow$ & LPIPS$\downarrow$ & SSIMAE$\downarrow$ \\
    \hline
    \multirow{4}{*}{\makecell[c]{1)}} & NSFF\cite{Li2021c} & 20.158 & 0.784 & 0.245 & 0.362 & 22.573 & 0.677 & 0.344 & 0.252 \\
    & DynNeRF\cite{Gao2021} & 20.106 & 0.768 & 0.256 & 0.318 & 25.647 & 0.846 & 0.172 & 0.205 \\
    & TiNeuVox\cite{Fang2022} & 20.632 & 0.692 & 0.247 & 0.423 & 24.506 & 0.702 & 0.266 & 0.397 \\
    & HexPlane\cite{Cao2023a} & 20.819 & 0.725 & 0.241 & 0.335 & 17.760 & 0.545 & 0.619 & 1.063 \\
    \hline
    \multirow{4}{*}{\makecell[c]{2)}} & NSFF\cite{Li2021c} & 20.247 & 0.783 & 0.246 & 0.365 & 20.550 & 0.624 & 0.419 & 0.290 \\
    & DynNeRF\cite{Gao2021} & 21.717 & 0.776 & 0.240 & 0.304 & 25.717 & 0.850 & 0.165 & 0.168 \\
    & TiNeuVox\cite{Fang2022} & 22.627 & 0.711 & 0.229 & 0.255 & 26.132 & 0.747 & 0.221 & 0.263 \\
    & HexPlane\cite{Cao2023a} & 15.885 & 0.525 & 0.542 & 1.197 & 19.819 & 0.574 & 0.495 & 1.257 \\
    \hline
    \multirow{5}{*}{\makecell[c]{3)}} & NSFF\cite{Li2021c} & 20.651 & 0.735 & 0.266 & 0.379 & 24.833 & 0.750 & 0.280 & 0.336 \\
    & DynNeRF\cite{Gao2021} & 20.026 & 0.760 & 0.264 & 0.342 & 25.529 & 0.842 & 0.178 & 0.199 \\
    & TiNeuVox\cite{Fang2022} & 21.480 & 0.710 & 0.241 & 0.260 & 25.302 & 0.743 & 0.234 & 0.233 \\
    & HexPlane\cite{Cao2023a} & 21.123 & 0.709 & 0.247 & 0.258 & 20.120 & 0.612 & 0.458 & 0.527 \\
    & \textbf{\nickname{}(Ours)} & \textbf{22.851} & \textbf{0.838} & \textbf{0.141} & 0.168 & \textbf{27.961} & \textbf{0.899} & \textbf{0.091} & \textbf{0.043} \\
    \hline
    2)+3) & Total-Recon\cite{Song2023} & 22.268 & 0.810 & 0.145 & \textbf{0.097} & 25.830 & 0.867 & 0.113 & 0.135 \\
    \hline
\end{tabular}
\end{scriptsize}
\label{tab:nvs_indoor} \vspace{-0.2cm}
\end{table*}

\begin{table*}[th] \vspace{-0.2cm}
\caption{Per-scene quantitative results for novel view synthesis on the Oxford Multimotion Dataset. The methods are trained with different depth supervision: 1) w/o depth, 2) w/ MiDaS depth, and 3) w/ per-object SfM depth.}
\centering
\begin{scriptsize}
\begin{tabular}{cl|ccc|ccc}
    \hline
    \multirow{2}{*}{} & & \multicolumn{3}{c}{swinging\_4\_translational} & \multicolumn{3}{c}{occlusion\_2\_translational} \\
    Depth Sup. & Method & PSNR$\uparrow$ & SSIM$\uparrow$ & LPIPS$\downarrow$ & PSNR$\uparrow$ & SSIM$\uparrow$ & LPIPS$\downarrow$ \\
    \hline
    \multirow{4}{*}{\makecell[c]{1)}} & NSFF\cite{Li2021c} & 15.460 & 0.523 & 0.301 & 17.159 & 0.640 & 0.259 \\
    & DynNeRF\cite{Gao2021} & 15.506 & 0.534 & 0.310 & 17.736 & 0.657 & 0.236 \\
    & TiNeuVox\cite{Fang2022} & 16.625 & 0.588 & 0.272 & 15.656 & 0.592 & 0.379 \\
    & HexPlane\cite{Cao2023a} & 16.595 & 0.580 & 0.208 & 16.759 & 0.637 & 0.253 \\
    \hline
    \multirow{4}{*}{\makecell[c]{2)}} & NSFF\cite{Li2021c} & 15.628 & 0.529 & 0.270 & 16.777 & 0.627 & 0.322 \\
    & DynNeRF\cite{Gao2021} & 15.909 & 0.543 & 0.251 & 17.232 & 0.634 & 0.283 \\
    & TiNeuVox\cite{Fang2022} & 17.320 & 0.613 & 0.200 & 18.503 & 0.645 & 0.220 \\
    & HexPlane\cite{Cao2023a} & 15.898 & 0.547 & 0.280 & 15.972 & 0.597 & 0.358 \\
    \hline
    \multirow{5}{*}{\makecell[c]{3)}} & NSFF\cite{Li2021c} & 16.590 & 0.566 & 0.307 & 17.373 & 0.628 & 0.254 \\
    & DynNeRF\cite{Gao2021} & 15.596 & 0.541 & 0.289 & 18.230 & 0.673 & 0.216 \\
    & TiNeuVox\cite{Fang2022} & 17.321 & \textbf{0.631} & 0.195 & 17.618 & 0.633 & 0.256 \\
    & HexPlane\cite{Cao2023a} & 16.263 & 0.558 & 0.209 & 17.078 & 0.643 & 0.219 \\
    & \textbf{\nickname{}(Ours)} & \textbf{18.248} & 0.619 & \textbf{0.160} & 19.551 & \textbf{0.686} & \textbf{0.169} \\
    \hline
    2)+3) & Total-Recon\cite{Song2023} & 16.029 & 0.546 & 0.184 & \textbf{19.597} & 0.682 & 0.194 \\
    \hline

    \multirow{2}{*}{} & & \multicolumn{3}{c}{swinging\_4\_unconstrained} & \multicolumn{3}{c}{occlusion\_2\_unconstrained} \\
    Depth Sup. & Method & PSNR$\uparrow$ & SSIM$\uparrow$ & LPIPS$\downarrow$ & PSNR$\uparrow$ & SSIM$\uparrow$ & LPIPS$\downarrow$ \\
    \hline
    \multirow{4}{*}{\makecell[c]{1)}} & NSFF\cite{Li2021c} & 16.140 & 0.578 & 0.221 & 17.987 & 0.723 & 0.215 \\
    & DynNeRF\cite{Gao2021} & 16.668 & 0.600 & 0.191 & 17.521 & 0.717 & 0.238 \\
    & TiNeuVox\cite{Fang2022} & 17.047 & 0.620 & 0.278 & 16.402 & 0.651 & 0.372 \\
    & HexPlane\cite{Cao2023a} & 16.883 & 0.607 & 0.189 & 18.097 & 0.701 & 0.235 \\
    \hline
    \multirow{4}{*}{\makecell[c]{2)}} & NSFF\cite{Li2021c} & 17.715 & 0.620 & 0.161 & 18.257 & 0.716 & 0.221 \\
    & DynNeRF\cite{Gao2021} & 15.688 & 0.559 & 0.303 & 17.256 & 0.751 & 0.197 \\
    & TiNeuVox\cite{Fang2022} & 17.889 & 0.661 & 0.192 & 20.320 & 0.753 & 0.177 \\
    & HexPlane\cite{Cao2023a} & 15.703 & 0.538 & 0.319 & 15.797 & 0.620 & 0.396 \\
    \hline
    \multirow{5}{*}{\makecell[c]{3)}} & NSFF\cite{Li2021c} & 17.521 & 0.627 & 0.177 & 16.887 & 0.644 & 0.241 \\
    & DynNeRF\cite{Gao2021} & 15.830 & 0.565 & 0.312 & 17.487 & 0.715 & 0.308 \\
    & TiNeuVox\cite{Fang2022} & 17.843 & \textbf{0.676} & 0.179 & 19.389 & 0.740 & 0.200 \\
    & HexPlane\cite{Cao2023a} & 16.346 & 0.582 & 0.207 & 18.861 & 0.742 & 0.176 \\
    & \textbf{\nickname{}(Ours)} & \textbf{19.010} & 0.663 & \textbf{0.154} & \textbf{21.874} & \textbf{0.813} & 0.138 \\
    \hline
    2)+3) & Total-Recon\cite{Song2023} & 16.555 & 0.595 & 0.182 & 21.142 & 0.795 & \textbf{0.132} \\
    \hline
\end{tabular}
\end{scriptsize}
\label{tab:nvs_multimotion}\vspace{-0.8cm}
\end{table*}

\begin{table*}[h]\vspace{-2.5cm}
\caption{Per-scene quantitative results for novel view synthesis on the NVIDIA Dynamic Scene Dataset. The methods are trained with different depth supervision: 1) w/o depth, 2) w/ MiDaS depth, and 3) w/ per-object SfM depth.}
\centering
\begin{scriptsize}
\begin{tabular}{cl|ccc|ccc|ccc}
    \hline
    \multirow{2}{*}{} & & \multicolumn{3}{c}{Balloon2} & \multicolumn{3}{c}{Skating} & \multicolumn{3}{c}{Truck} \\
    Depth Sup. & Method & PSNR$\uparrow$ & SSIM$\uparrow$ & LPIPS$\downarrow$ & PSNR$\uparrow$ & SSIM$\uparrow$ & LPIPS$\downarrow$ & PSNR$\uparrow$ & SSIM$\uparrow$ & LPIPS$\downarrow$ \\
    \hline
    \multirow{4}{*}{\makecell[c]{1)}} & NSFF\cite{Li2021c} & 22.111 & 0.731 & 0.141 & 24.115 & 0.799 & 0.191 & 19.073 & 0.476 & 0.356 \\
    & DynNeRF\cite{Gao2021} & 24.391 & 0.783 & 0.114 & 26.973 & 0.920 & 0.073 & 25.750 & 0.779 & 0.164 \\
    & TiNeuVox\cite{Fang2022} & 21.702 & 0.522 & 0.263 & 23.371 & 0.700 & 0.275 & 23.693 & 0.633 & 0.249 \\
    & HexPlane\cite{Cao2023a} & 20.192 & 0.536 & 0.214 & 21.316 & 0.661 & 0.302 & 18.998 & 0.469 & 0.342 \\
    \hline
    \multirow{4}{*}{\makecell[c]{2)}} & NSFF\cite{Li2021c} & 25.887 & 0.838 & 0.072 & 29.457 & 0.927 & 0.057 & 27.034 & 0.817 & 0.097 \\
    & DynNeRF\cite{Gao2021} & 27.059 & \textbf{0.860} & \textbf{0.049} & 32.550 & \textbf{0.952} & \textbf{0.033} & \textbf{28.748} & 0.873 & 0.079 \\
    & TiNeuVox\cite{Fang2022} & 20.580 & 0.480 & 0.246 & 24.812 & 0.757 & 0.148 & 23.694 & 0.625 & 0.186 \\
    & HexPlane\cite{Cao2023a} & 17.538 & 0.317 & 0.328 & 21.459 & 0.621 & 0.287 & 18.938 & 0.474 & 0.386 \\
    \hline
    \multirow{5}{*}{\makecell[c]{3)}} & NSFF\cite{Li2021c} & 24.288 & 0.767 & 0.137 & 25.469 & 0.812 & 0.199 & 21.501 & 0.621 & 0.245 \\
    & DynNeRF\cite{Gao2021} & 21.969 & 0.658 & 0.202 & 26.552 & 0.880 & 0.163 & 24.973 & 0.775 & 0.164 \\
    & TiNeuVox\cite{Fang2022} & 19.837 & 0.427 & 0.275 & 24.467 & 0.705 & 0.160 & 23.770 & 0.642 & 0.210 \\
    & HexPlane\cite{Cao2023a} & 20.391 & 0.627 & 0.179 & 24.213 & 0.804 & 0.146 & 25.057 & 0.728 & 0.126 \\
    & \textbf{\nickname{}(Ours)} & \textbf{26.764} & 0.848 & 0.050 & \textbf{33.461} & 0.951 & 0.036 & 28.538 & \textbf{0.877} & \textbf{0.073}\\
    \hline
    2)+3) & Total-Recon\cite{Song2023} & 25.413 & 0.828 & 0.062 & 29.426 & 0.936 & 0.039 & 28.626 & 0.876 & 0.075 \\
    \hline
\end{tabular}
\end{scriptsize}
\label{tab:nvs_dynamicscene}
\end{table*}

